\pdfoutput=1

\documentclass[11pt]{article}

\usepackage[]{EMNLP2023}

\usepackage{times}
\usepackage{latexsym}

\usepackage[T1]{fontenc}

\usepackage[utf8]{inputenc}

\usepackage{microtype}

\usepackage{inconsolata}

\usepackage{xspace}
\usepackage{graphicx}
\usepackage{booktabs}
\usepackage{multirow}
\usepackage{amssymb}
\usepackage{xcolor}
\usepackage{subcaption}
\usepackage{nameref}
\usepackage{enumitem}
\usepackage{empheq, bm}
\usepackage{listings}

\usepackage{todonotes}
\usepackage{menukeys}
\usepackage[all]{nowidow}
 
\definecolor{codegreen}{rgb}{0,0.6,0}
\definecolor{codegray}{rgb}{0.5,0.5,0.5}
\definecolor{codepurple}{rgb}{0.58,0,0.82}
\definecolor{backcolour}{rgb}{0.95,0.95,0.92}
 
\lstdefinestyle{mystyle}{
    backgroundcolor=\color{backcolour},   
    commentstyle=\color{codegreen},
    keywordstyle=\color{magenta},
    numberstyle=\tiny\color{codegray},
    stringstyle=\color{codepurple},
    basicstyle=\ttfamily\scriptsize,
    breakatwhitespace=false,         
    breaklines=true,                 
    captionpos=b,                    
    keepspaces=true,                 
    numbers=left,                    
    numbersep=5pt,                  
    showspaces=false,                
    showstringspaces=false,
    showtabs=false,                  
    tabsize=2
}
 
\usepackage{lipsum}
\usepackage{comment}

\definecolor{backcolour}{rgb}{0.95,0.95,0.92}

\title{ConvLab-3: A Flexible Dialogue System Toolkit \\Based on a Unified Data Format}

\author{Qi Zhu$^{*1}$ \quad Christian Geishauser$^{*2}$ \quad Hsien-chin Lin$^{*2}$ \quad Carel van Niekerk$^{*2}$ \\ \textbf{Baolin Peng$^3$} \quad \textbf{Zheng Zhang$^1$} \quad 
\textbf{Shutong Feng$^2$} \quad \textbf{Michael Heck$^2$} \quad \textbf{Nurul Lubis$^2$} \\ \textbf{Dazhen Wan$^1$} \quad \textbf{Xiaochen Zhu$^4$} \quad \textbf{Jianfeng Gao$^3$} \quad \textbf{Milica Ga\v{s}i\'c$^{\dagger2}$}
\quad \textbf{Minlie Huang$^{\dagger1}$}\\
  $^1$Tsinghua University, Beijing, China \\
  $^2$Heinrich Heine University Düsseldorf, Düsseldorf, Germany \\
  $^3$Microsoft Research, Redmond, USA \\
  $^4$University of Cambridge, Cambridge, England \\
  $^1${\small \tt \{zhu-q18,wandz19\}@mails.tsinghua.edu.cn \quad \{z-zhang,aihuang\}@tsinghua.edu.cn} \\
  $^2${\small \tt \{geishaus,linh,niekerk,heckmi,lubis,gasic\}@hhu.de} \\
  $^3${\small \tt \{bapeng,jfgao\}@microsoft.com} \quad
  $^4${\small \tt xz479@cam.ac.uk}}

\begin{document}
\maketitle
\begin{abstract}
Task-oriented dialogue (TOD) systems function as digital assistants, guiding users through various tasks such as booking flights or finding restaurants. Existing toolkits
for building TOD systems 
often fall short of in delivering comprehensive arrays of data, models, and experimental environments with a user-friendly experience. 
We introduce ConvLab-3: a multifaceted dialogue system toolkit crafted to bridge this gap. Our unified data format simplifies the integration of diverse datasets and models, significantly reducing complexity and cost for studying generalization and transfer. Enhanced with robust reinforcement learning (RL) tools, featuring a streamlined training process, in-depth evaluation tools, and a selection of user simulators, ConvLab-3 supports the rapid development and evaluation of robust dialogue policies. Through an extensive study, we demonstrate the efficacy of transfer learning and RL and showcase that ConvLab-3 is not only a powerful tool for seasoned researchers but also an accessible platform for newcomers\footnote{ConvLab-3 is publicly available at \url{https://github.com/ConvLab/ConvLab-3} under Apache License 2.0. The demonstrative video accompanying this paper is available at \url{https://youtu.be/t6HVTJCeGLo}.}.
\end{abstract}

\section{Introduction}

\def\thefootnote{*}\footnotetext{These authors contributed equally to this work.} \def\thefootnote{$\dagger$}\footnotetext{These authors share the senior authorship of this work.} Task-oriented dialogue~(TOD) systems converse with their users in natural language to help them fulfil a task, such as booking a flight or finding a restaurant. Unlike chit-chat dialogues, a critical aspect of these systems is that they are grounded in an ontology that contains domains, slots, and values which describe the dialogue task, i.e. user goal, as well as including domain-specific databases.

\begin{figure*}[ht!]
    \centering
    \includegraphics[width=\linewidth]{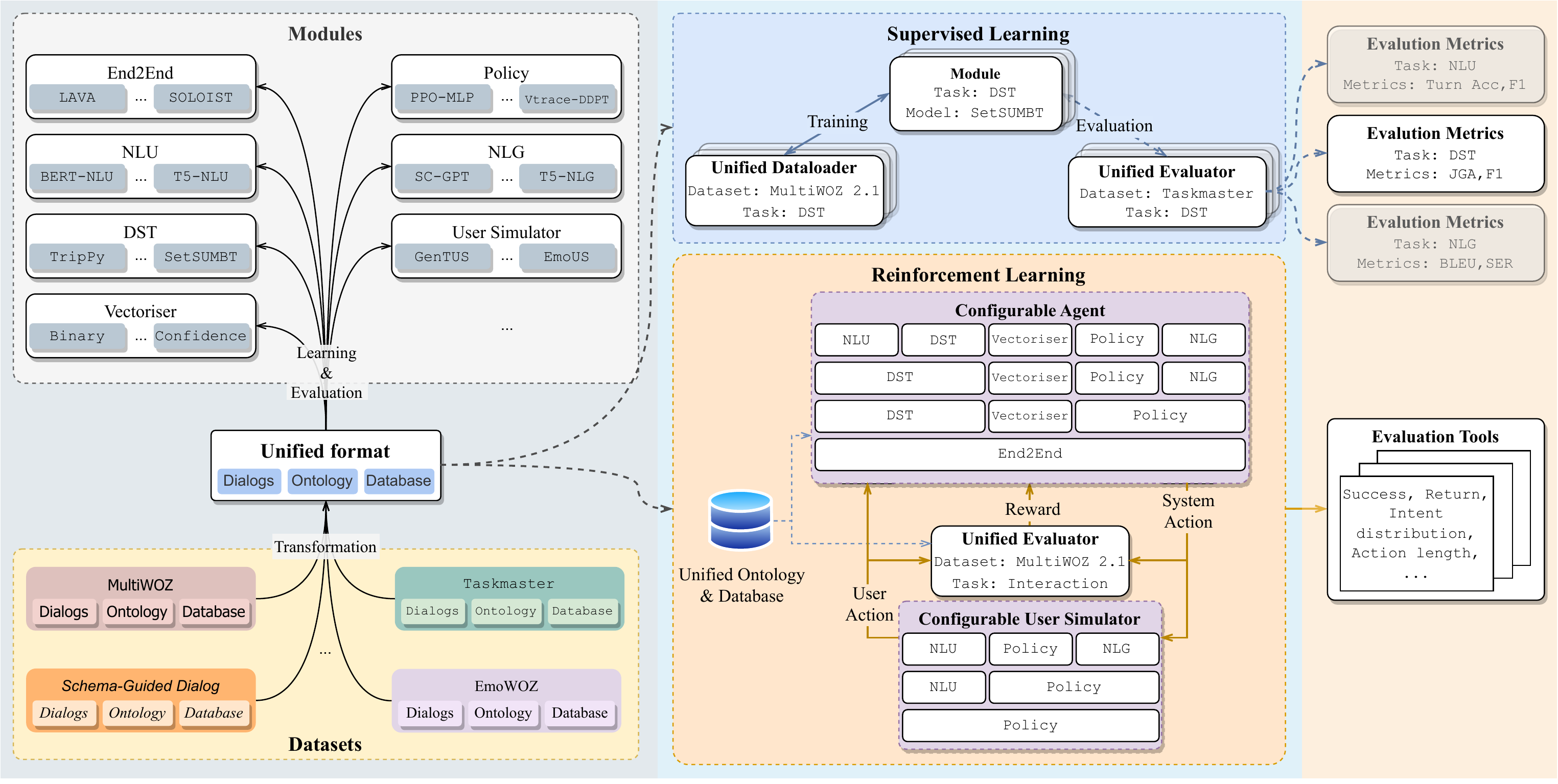}
    \caption{ConvLab-3: The unified format serves as a bridge, connecting diverse datasets and dialogue models. It streamlines the integration of various TOD modules, including supervised learning, evaluation, and a wide array of essential evaluation metrics, thanks to the unified data loader and evaluator. These modules can be incorporated, either in the agent or user simulator, through a configuration file, defining the environment for interactive evaluation and reinforcement learning.}
    \label{fig:framework}
\end{figure*}

There are two distinct capabilities that TOD systems need to exhibit. They need to \emph{track} the state of the dialogue and based on that \emph{decide} on the next action to take in order to steer the conversation towards fulfilling the user's goal~\citep{young2007his}. The architecture of TOD systems typically adopts a modular approach, often encompassing components like dialogue state trackers and policies, and may include language understanding or generation units, as depicted in Figure~\ref{fig:framework}. The complexity of a TOD system necessitates a toolkit with advanced, easily integrable modules allowing for straightforward training, evaluation, and combination.

The vast amount of possible user behaviours and tasks that a TOD system might assist with necessitates the study of generalization and transfer towards new users and datasets. While many datasets for studying task-oriented dialogue have been proposed ~\citep{wen-etal-2016-conditional,mrksic-etal-2017-neural,byrne-etal-2019-taskmaster,eric-etal-2020-multiwoz,rastogi2020sgd,zhu-etal-2020-crosswoz,feng-etal-2022-emowoz}, the various dialogue, ontology and database formats hinder researchers from validating their models on unseen data. In this work we propose a unified format to bridge the gap between different TOD datasets and models and provide a unified training and evaluation framework that accelerates the study of generalization capabilities. Once a dataset is transformed into the unified format, it can be immediately used by supported models. Similarly, once a model supports the unified format, it can access all supported datasets. This feature reduces the cost of adapting $M$ models to $N$ datasets from $M \times N$ to $M + N$.

The dialogue policy, as the decision-making component of a TOD system, is pivotal to the success or failure of a dialogue task. It is typically optimized using reinforcement learning (RL), necessitating additional components such as algorithms, evaluation tools, and user simulators. Realistic user simulators are essential for conducting interactive evaluations and tests against varied user behaviours, in order to accurately mirror real-world scenarios. ConvLab-3 streamlines RL-based development and assessment of dialogue policies. We achieve this by offering a configurable RL environment, evaluation tools for thorough insights, and multiple user simulators to explore generalization capabilities towards new user behaviours, as depicted in Figure~\ref{fig:framework}.

ConvLab-3 is especially useful for practitioners seeking to construct a dialogue system without extensive expertise. Additionally, it provides a fast, convenient, and dependable platform for both novice and experienced researchers to conduct experiments. In particular, it enables: (1) researchers to perform experiments across a variety of datasets, (2) developers to construct an dialogue system using custom datasets, and (3) community contributors to consistently add models and datasets. In summary, our contributions are:

\begin{itemize}
    \item A unified data format which allows for easy generalisation and transfer learning experiments across different datasets.
    \item A convenient RL framework and access to different user simulators, accelerating the development and evaluation of dialogue policies.
    \item Providing a broad collection of compatible datasets and state-of-the-art models.
\end{itemize}

\section{Related work}
\label{sec:relatedwork}


While Rasa \cite{bocklisch2017rasa}, NeMo \cite{kuchaiev2019nemo} and DialogueStudio \cite{zhang2023dialogstudio} provide unified data formats, they do not have RL tools or user simulators for interactive training and evaluation of dialogue systems. ParlAI \cite{miller-etal-2017-parlai} includes a \emph{reward} attribute in their unified format, but without accessible RL tools. PyDial \cite{ultes-etal-2017-pydial} and the predecessors of ConvLab-3 \cite{lee-etal-2019-convlab,zhu-etal-2020-convlab} provide reinforcement learning toolkits, however they lack a unified format and thus the possibility to study generalization across datasets. Moreover, PyDial and previous versions of ConvLab do not provide multiple data-driven user simulators and their training evaluation provides no tools for in-depth analysis. In addition, none of the above toolkits provide a sufficient set of state-of-the-art models for the different components in a TOD system.

\begin{table}[ht!]
\small
\centering
\resizebox{\columnwidth}{!}{%
\setlength{\tabcolsep}{0.9mm}{
\begin{tabular}{l|cccccc}
\toprule
\multirow{2}{*}{Dataset}  & \multicolumn{6}{c}{Dataset Annotations} \\
& Goal & DA-U & DA-S & State & API& Database \\
\midrule
Camrest~\citeyearpar{wen-etal-2016-conditional} & \checkmark & \checkmark & \checkmark & \checkmark & & \checkmark \\
WOZ $2.0$~\citeyearpar{mrksic-etal-2017-neural}& & \checkmark & & \checkmark & & \\
KVRET~\citeyearpar{eric-etal-2017-key} & & \checkmark & & \checkmark & \checkmark \\
DailyDialog~\citeyearpar{li-etal-2017-dailydialog} & & \checkmark & & & & \\
Taskmaster-$1$~\citeyearpar{byrne-etal-2019-taskmaster}  & & \checkmark & \checkmark & \checkmark & & \\
Taskmaster-$2$~\citeyearpar{byrne-etal-2019-taskmaster}  & & \checkmark & \checkmark & \checkmark & & \\
MultiWOZ $2.1$~\citeyearpar{eric-etal-2020-multiwoz} & \checkmark & \checkmark & \checkmark & \checkmark & & \checkmark \\
SGD~\citeyearpar{rastogi2020sgd} & & \checkmark & \checkmark & \checkmark & \checkmark & \\
MetaLWOZ~\citeyearpar{li2020metalwoz} &  \checkmark & & & & & \\
CrossWOZ~\citeyearpar{zhu-etal-2020-crosswoz} & \checkmark & \checkmark & \checkmark & \checkmark & \checkmark & \checkmark \\
Taskmaster-$3$~\citeyearpar{byrne-etal-2021-tickettalk}  & & \checkmark & \checkmark & \checkmark & \checkmark & \\
EmoWOZ~\citeyearpar{feng-etal-2022-emowoz} & \checkmark & \checkmark & \checkmark & \checkmark & & \checkmark \\
\bottomrule
\end{tabular}
}%
}
\caption{Annotations of current unified datasets. DA-U/DA-S is dialogue acts annotation of user/system.}
\label{tab:datasets-annotation}
\end{table}

\section{Unified Format}
\label{unified-data-format-section}

In our unified format, a dataset consists of (1) an \textbf{ontology} that defines the annotation schema, (2) \textbf{dialogues} with transformed annotations, and (3) a \textbf{database} that links to external knowledge sources (see Figure~\ref{fig:framework}).

Typically converting the formats of different datasets is not straightforward, hindering format adaptation of existing and new corpora. However, in ConvLab-3 we provide detailed guidelines and scripts that make the process of format adaptation straightforward and error-free. ConvLab-3 offers a large number of datasets in the unified format as shown in Table~\ref{tab:datasets-annotation}, whilst also simplifying the process of adding new datasets.

Moreover, as shown in Listing \ref{listing:util}, we provide utility functions to process the unified datasets, such as delexicalization, splitting data for few-shot learning, and loading data for specific tasks. Based on the unified format, evaluations of common tasks across models and corpora are standardized, which facilitates comparability. 
More details of already supported datasets and tasks can be found in Appendix~\ref{sec:annotation} and \ref{sec:tasks}, respectively.

\subsection{Ontology}
Following \citet{budzianowski-etal-2018-multiwoz} and \citet{rastogi2020sgd}, an ontology consists of:
(1) Domains and their slots in a hierarchical format. Each slot has a Boolean flag indicating whether it is a categorical slot (whose value set is fixed).
(2) All possible intents in dialogue acts.
(3) Possible dialogue acts appearing in the dialogues. Each act is comprised of intent, domain, slot, and speaker (i.e., system or user).
(4) Template dialogue state.
We also provide a natural language description, if available, for each domain, slot, and intent to facilitate few-shot learning \cite{mifei2022cins} and domain transfer \cite{lin-etal-2021-leveraging}.

\begin{figure}
\begin{lstlisting}[language=Python, numbers=none, caption={Example usage of unified datasets.}, label={listing:util}]
from convlab.util import *

dataset_name = "multiwoz21"
# load dataset: a dict maps data_split to dialogues
dataset = load_dataset(dataset_name)
# load dataset in a predefined order with a custom 
# split ratio for reproducible few-shot experiments
dataset = load_dataset(dataset_name, \ 
                       split2ratio={"train": 0.01})

# load ontology and database similarly
ontology = load_ontology(dataset_name)
database = load_database(dataset_name)
# query the database with domain and state
state = {"hotel": {"area": "east", \
                   "price range": "moderate"}}
res = database.query("hotel", state, topk=3)

# Example functions based on the unified format
# load the user turns in the test set for NLU task
nlu_data = load_nlu_data(dataset, "test", "user")
# dataset-agnostic delexicalization
dataset, delex_vocab = create_delex_data(dataset)
\end{lstlisting}
\vspace{-0.7cm}
\end{figure}

\subsection{Dialogues}

We unify the format of dialogue annotations included in many datasets and commonly used by dialogue models while keeping the original format of annotations that only appear in specific datasets. 
As we integrate more datasets in the future, we will expand the unified format to include more common annotations.

For a dialogue in the unified format, dialogue-level information includes the dataset name, data split (training or test), unique dialogue ID, involved domains, user goal, etc.
Following MultiWOZ \cite{budzianowski-etal-2018-multiwoz}, a user goal has informable slot-value pairs, requestable slots, and a natural language instruction summarizing the goal.

Turn-level information includes speaker, utterance, dialogue acts, state, database result, etc (see Appendix~\ref{appendix:section:multiwoz-example} for an example).
Each dialogue act is a list of tuples, each tuple consisting of intent, domain, slot, and value.
According to the value, we divide dialogue acts into three groups: 
(1) \textbf{categorical} for slots whose value set is predefined in the ontology (e.g., inform the weekday of a flight).
(2) \textbf{non-categorical} for slots whose values can not be enumerated (e.g., inform the address of a hotel).
(3) \textbf{binary} for intents without actual values (e.g., request the address of a hotel).
The state is initialized by the template state as defined by the ontology and updated during the conversation, containing slot-value pairs of involved domains.
A database result is a list of entities retrieved from the database or other knowledge sources.
We list common annotations included in the unified data format and the tasks they support in Appendix~\ref{sec:tasks}.
Other dataset-specific annotations are retained in their original formats.

\subsection{Database/API Interface}

To unify the interaction with different types of databases, we define a \texttt{BaseDatabase} class that has an abstract \texttt{query} function to be customized. The \texttt{query} function takes the current domain, dialogue state, and other custom arguments as input and returns a list of top-k candidate entities. By inheriting \texttt{BaseDatabase} and overriding the \texttt{query} function, we can easily access different databases/APIs and retrieve the result with a unified format.

\subsection{Evaluation}

To provide a comparable evaluation setup for all TOD tasks supported by the unified format, we provide unified evaluation scripts. These scripts include commonly used metrics such as: turn accuracy (ACC) and dialogue act F1 score for natural language understanding~(NLU)~\citep{zhu-etal-2020-convlab}, joint goal accuracy (JGA) and slot F1 score for dialogue state tracking~(DST)~\citep{li2021dstc9}, BLEU and slot error rate (SER) for natural language generation~(NLG)~\citep{wen-etal-2015-semantically}, BLEU and Combined score (Comb.) for End2End dialogue modeeling~\citep{mehri-etal-2019-structured}, turn accuracy, slot-value F1 score and SER for user simulators~\citep{lin-etal-2021-domain,lin-etal-2022-gentus}.

\section{Integrated Models}
\label{sec:newmodels}



Convlab-3 provides a wide array of standard and state-of-the-art models covering all modules in a TOD system. This allows straightforward plug-and-play experimentation when developing a specific module, as well as building TOD systems easily on custom datasets. A model is considered integrated once it implements the corresponding module interface and supports processing datasets in the unified format.

Besides existing models in ConvLab-2 \cite{zhu-etal-2020-convlab}, we integrate new transformer-based models supporting the unified data format, including SetSUMBT \cite{van-niekerk-etal-2021-uncertainty} and TripPy \cite{heck-etal-2020-trippy} for dialogue state tracking (DST), DDPT \cite{geishauser-etal-2022-dynamic} and LAVA \cite{lubis-etal-2020-lava} for policy learning, SC-GPT \cite{peng-etal-2020-shot} for natural language generation (NLG), and SOLOIST \cite{peng-etal-2021-soloist} with T5 as backbone model \cite{peng2022godel} for end-to-end modeling (End2End).
We also integrate multiple powerful data-driven user simulators (US): TUS \cite{lin-etal-2021-domain} that outputs user dialogue acts, GenTUS \cite{lin-etal-2022-gentus} that outputs both user dialogue acts and response, and EmoUS \cite{lin23-emous} that additionally outputs emotions. 

In addition, we apply text-generation models to solve the tasks of TOD modules (see Appendix~\ref{appendix:section:serialization}. We provide a range of models built upon T5~\citep{raffel2020t5}, covering NLU, DST, NLG, etc. We also provide an interface to instruct large language models (LLMs) such as ChatGPT and LLaMa \cite{llama2} to serve as different modules such as user simulators, NLU, DST, NLG, etc. See~\citet{heck-etal-2023-chatgpt} for an example of how ChatGPT can be instructed to serve as a DST model. All integrated models are shown in Appendix~\ref{sec:tasks}.

\section{Reinforcement Learning Toolkit}
\label{sec:rltools}


The difficulty of building a comprehensive TOD toolkit lies in the fact that it needs to support not only supervised but also reinforcement learning. As shown in Figure \ref{fig:framework}, this includes functionalities to build configurable agents and user simulators consisting of different modules, an evaluator to provide reward signals, and analysis tools to evaluate the training process and RL algorithms. ConvLab-3 supports the straightforward combination of components with an easy-to-use configuration file, including the definition of the interactive environment given by the choice of user policy and its components, see Appendix \ref{appendix:section:config-file} for an example. 

The dialogue policy module obtains the semantic information of the DST (and NLU) as input and produces a list of atomic actions $[({\tt domain_1,intent_1,slot_1}), ...]$ as output, e.g. $[({\tt hotel,inform,phone}), ({\tt hotel,inform,}$ ${\tt addr})]$, which results in a large action space due to the high number of possible atomic actions and their combinations. As the input is on semantic level while the policy network expects vectorized input, ConvLab-3 provides a \texttt{Vectoriser} class that acts as communication module between semantic and vector representation. We treat the Vectoriser as an additional pipeline module, which allows straightforward investigation of different vectorization strategies in a plug-and-play fashion. Moreover, policy networks can be used off-the-shelf while only the Vectoriser needs to be adapted. ConvLab-3 provides a base Vectoriser class that can be easily adapted, as well as common vectorization strategies. In addition, we add the possibility for masking certain actions as in\citet{ultes-etal-2017-pydial}. This allows controllability of the policy output and facilitates learning during RL due to reduction of the large action space. Moreover, in addition to the on-policy RL algorithms REINFORCE \cite{sutton99} and PPO \cite{ppo}, which are already implemented in ConvLab-2, we provide the state-of-the-art continual RL model DDPT together with state-of-the-art algorithms VTRACE \cite{espeholt18-impala} and CLEAR \cite{rolnick19} for off-policy~\citep{Sutton18} and continual RL \cite{towardsCRL}, respectively. 

\subsection{Evaluation Tools}

Understanding the policy behaviour allows researchers to fine-tune their algorithm or reward model in an informed manner to improve performance. The analysis of policy behaviour can be done by studying 1) the efficiency of actions, i.e. how many atomic actions are taken in a turn, 2) how the selected intents are distributed in a turn, 3) actual dialogue interactions. The average number of atomic actions is an important indicator of information overload, which a user simulator can handle well in contrast to humans. The intent distribution reveals policy preferences and possible exploitations of imperfect user simulators. 

ConvLab-3 is the first toolkit to provide these set of measures and evaluation tools together with the common measurements of task success, return and average number of turns. Moreover, actual dialogues can be observed for in-depth evaluation. 


\section{Supervised Learning Experiments}
\label{sec:supervisedlearning}

Conducting supervised learning experiments on multiple TOD datasets is convenient with the unified data format. We believe this feature will encourage researchers to build general dialogue models that perform well on various data as well as to investigate knowledge transfer. In these experiments, we demonstrate the ease of evaluating a model's knowledge transfer abilities using our unified format. Initially, we pre-train all models on the Schema-Guided Dialogue (SGD)~\citep{rastogi2020sgd} and Taskmaster-$1\&2\&3$~\citep{byrne-etal-2019-taskmaster,byrne-etal-2021-tickettalk} datasets jointly. These models are then fine-tuned on MultiWOZ $2.1$~\citep{eric-etal-2021-multi} in full-data or low-resource settings. To configure these different training setups, one only needs to make a few changes to the unified dataloader parameters, as depicted in Listing~\ref{listing:util}. For low-resource fine-tuning, we set the data ratios of both training and validation set to $1\%$ and $10\%$.

\begin{table}[t]
\small
\centering
\setlength{\tabcolsep}{0.5mm}{
\begin{tabular}{@{}lcc|cc|cc@{}}
\toprule
                        & \multicolumn{6}{c}{MultiWOZ 2.1}                                              \\
                        & \multicolumn{2}{c}{1\%} & \multicolumn{2}{c}{10\%} & \multicolumn{2}{c}{100\%}\\ \cmidrule{2-7}
\textbf{DST}            & JGA $\uparrow$       & Slot F1 $\uparrow$   & JGA $\uparrow$       & Slot F1 $\uparrow$   & JGA $\uparrow$       & Slot F1 $\uparrow$           \\ \midrule
\multirow{2}{*}{T5-DST} & $14.5$      & $68.5$      & $35.5$      & $84.8$      & $52.6$      & $91.9$              \\
                        & $\bm{22.9}$      & $\bm{74.9}$      & $\bm{41.2}$      & $\bm{87.1}$      & $53.1$      & $92.0$              \\ \midrule
\multirow{2}{*}{SetSUMBT}& $7.8$         & $41.8$          & $37.0$          & $84.4$          & $50.3$          & $90.8$                  \\
                        & $\bm{22.7}$          & $\bm{77.2}$          & $\bm{43.8}$          & $\bm{88.2}$          & $50.7$          & $91.2$                  \\ \midrule \midrule
\textbf{NLG}            & SER $\downarrow$      & BLEU $\uparrow$      & SER $\downarrow$      & BLEU $\uparrow$      & SER $\downarrow$      & BLEU $\uparrow$              \\ \midrule
\multirow{2}{*}{T5-NLG} & $19.0$      & $20.2$      & $6.9$       & $31.3$      & $3.7$       & $35.8$              \\
                        & $\bm{9.8}$       & $\bm{25.8}$      & $\bm{5.5}$       & $\bm{32.9}$      & $3.5$       & $35.8$              \\ \midrule
\multirow{2}{*}{SC-GPT} &      $27.3$     &   $14.1$    &    $11.2$       &    $28.4$       &     $4.8$      &   $33.6$                \\
                        &  $\bm{9.5}$     &      $\bm{26.3}$    &  $\bm{6.9}$         &     $28.6$      &   $5.3$        &     $32.1$               \\  \midrule \midrule
\textbf{End2End}        & Comb.     & BLEU      & Comb.     & BLEU      & Comb.     & BLEU                       \\ \midrule
\multirow{2}{*}{SOLOIST}& $19.8$      & $0.4$       & $48.0$      & $10.0$      & $67.0$      & $16.8$              \\
                        & $\bm{42.2}$      & $\bm{10.4}$       & $\bm{62.0}$      & $\bm{15.9}$      & $\bm{71.4}$      & $17.5$              \\ \bottomrule
\end{tabular}
}
\caption{Comparison between models without pre-training (1st row) and with pre-training (2nd row) in both the low-resource and full-data settings.}
\vspace{-1em}
\label{tab:transfer}
\end{table}

In the low-resource setting, we observe that pre-training is beneficial, as evidenced in Table~\ref{tab:transfer}. Specifically for the end-to-end model SOLOIST, pre-training also proves advantageous in the full-data setting. This may be attributed to the increased complexity of the end-to-end modeling task. These findings emphasize that transfer learning can be successfully implemented in ConvLab-3 in a straightforward way. This enables: (1) developers to leverage knowledge from existing datasets for application in smaller, custom settings; (2) newcomers to explore the capabilities of various models; and (3) experienced researchers to evaluate the generalisability of their proposed methods, as well as to compare them to the available state-of-the-art benchmarks. For an example of joint training across multiple datasets and retrieval based data augmentation, see Appendix~\ref{sec:app_joint_training} and~\ref{app:retrieval-augmentation}.

\begin{figure*}[t!]
  \begin{subfigure}[t]{0.32\textwidth}
    \includegraphics[scale=0.35]{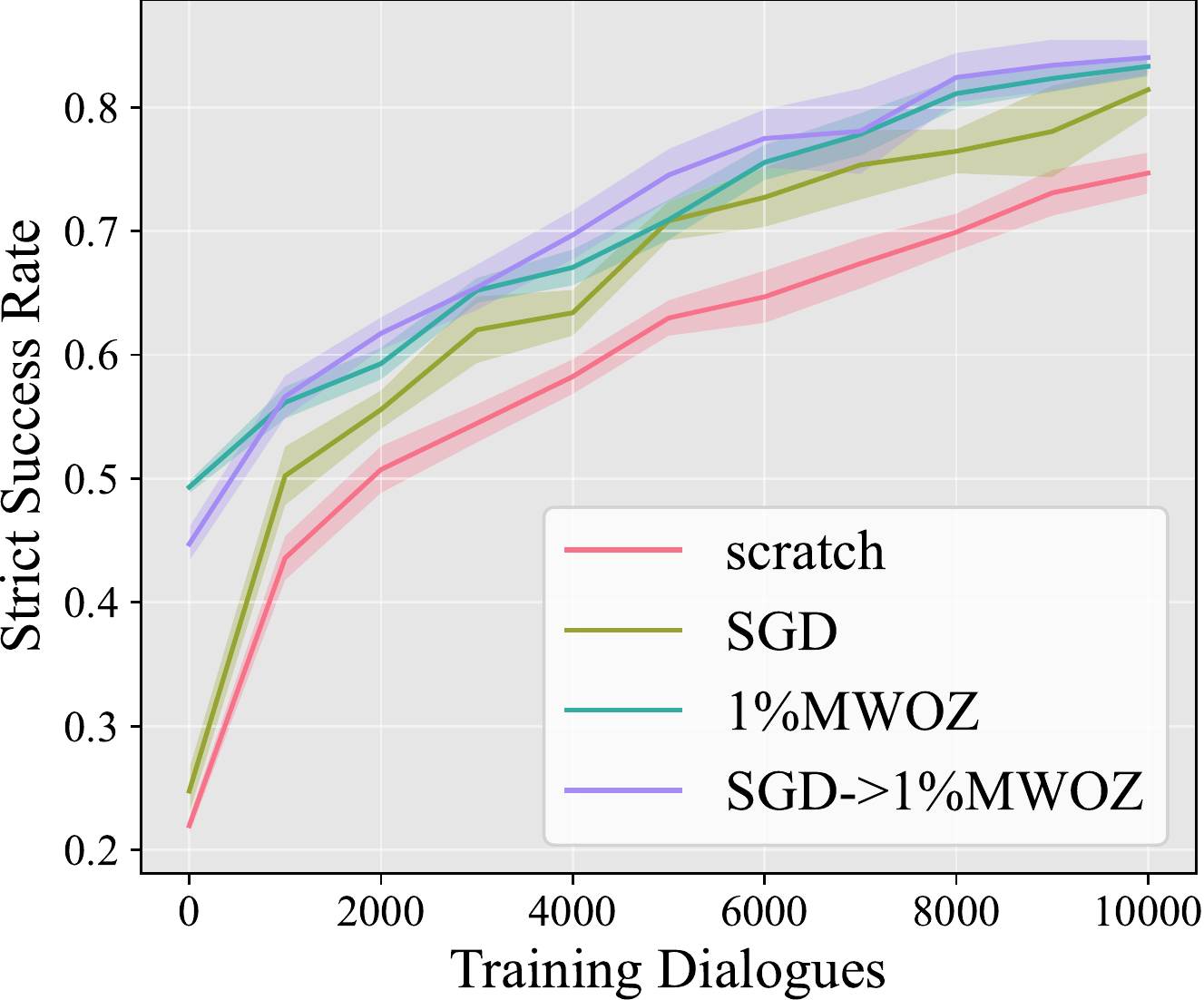}
    \caption{Strict success rate}
  \end{subfigure}
  \hspace{0.01\textwidth}%
  \begin{subfigure}[t]{0.32\textwidth}
    \includegraphics[scale=0.35]{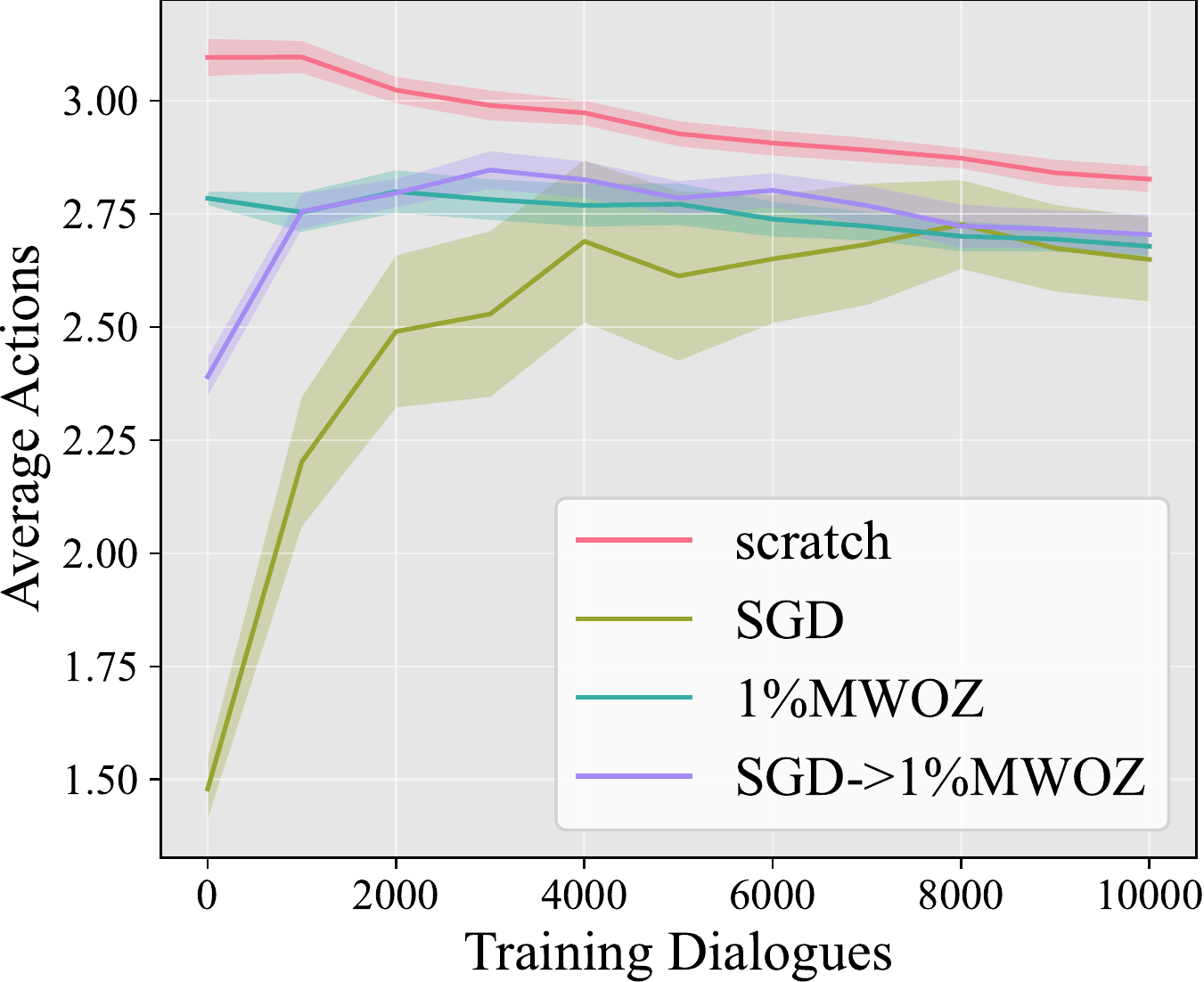}
    \caption{Average number of actions per turn}
  \end{subfigure}
  \hspace{0.01\textwidth}%
  \begin{subfigure}[t]{0.32\textwidth}
    \includegraphics[scale=0.35]{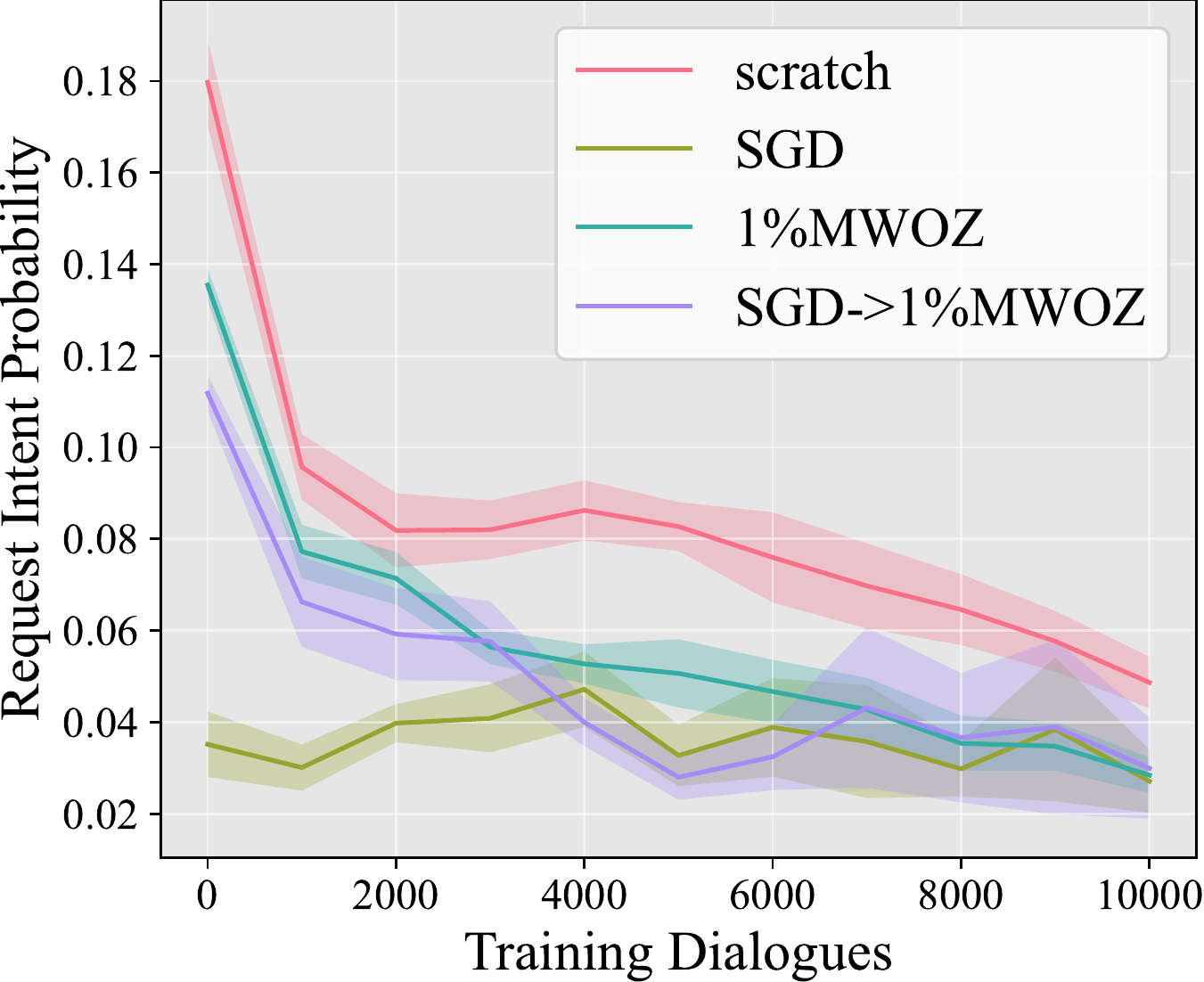}
    \caption{Probability of the request intent}
  \end{subfigure}

\caption{Pre-training then RL training experiments with the DDPT model in interaction with the rule-based simulator. Shaded regions show standard error. Each model is evaluated on 9 different seeds.}
\label{pretrain-rl}
\end{figure*}

\section{Reinforcement Learning Experiments} \label{sec:rl-experiments}

ConvLab-3 supports a convenient way to run RL training and evaluation supported by the unified format and availability of multiple user simulators. To showcase this, we run transfer learning experiments as well as experiments with multiple user simulators.

\subsection{Transfer Learning}

We utilize the DDPT policy model with VTRACE as algorithm and consider four different data set scenarios for supervised pre-training: (1) \textbf{scratch} that does not use pre-training, (2) \textbf{SGD} that pre-trains on SGD, (3) \textbf{1\% MWOZ} that pre-trains on 1\% of MultiWOZ data, and (4) \textbf{SGD->1\%MWOZ} that pre-trains on SGD data and afterwards 1\% of MultiWOZ data. The experiments are conducted on the semantic level, leveraging the rule-based dialogue state tracker and the rule-based user simulator \cite{schatzmann-etal-2007-agenda} of ConvLab-3.

The results, depicted in Figure \ref{pretrain-rl}, show a similar trend for all models and metrics. Nevertheless, Figure \ref{pretrain-rl}(a) reveals that pre-training on SGD does not yield an advantage for the starting performance, compared to training from scratch, while it leads to better results by the end of training. Moreover, the number of actions taken in a turn and the probability of taking a request intent, shown in Figure \ref{pretrain-rl}(b) and (c), is initially much lower for the model trained on SGD only. This indicates that the behaviour learned from SGD differs significantly from the behaviour on MultiWOZ. Refer to Appendix \ref{app:rl-experiments-chris} for more results and experiments. Our unique evaluation tools thus provides essential insights into both metrics and the behaviour of the agent.

\begin{table}[t]
\small
\centering
\begin{tabular}{@{}lccc@{}}
\toprule
\multirow{2}{*}{\begin{tabular}[c]{@{}l@{}}US for \\ training\end{tabular}} & \multicolumn{3}{c}{US for testing} \\
 & ABUS & TUS & GenTUS \\ \midrule
ABUS & $\bm{0.93}$ & $0.71$ & $0.56$ \\
TUS & $0.87$ & $0.79$ & $0.59$ \\
GenTUS & $0.89$ & $\bm{0.86}$ & $\bm{0.63}$ \\ \bottomrule
\end{tabular}
\caption{The strict success rates of PPO-MLP policies trained on ABUS, TUS, and GenTUS when evaluated with various user simulators.}
\vspace{-1em}
\label{tab:cross-model}
\end{table}

\subsection{Evaluation across Different User Simulators}
\label{sec:user-simulators}

To enable a policy to generalize to diverse user behaviour, it's crucial to train and evaluate policy models across various user simulators. ConvLab-3 not only offers state-of-the-art data-driven user simulation models but also a configurable interactive environment for evaluation and reinforcement learning, as illustrated in Figure~\ref{fig:framework}. In these experiments, we utilise a multi-layer perceptron (MLP) policy trained with the PPO algorithm, using three distinct user simulators: ABUS \cite{schatzmann07-sim}, TUS, and GenTUS. We then evaluate the resulting policies using each of these simulators. 

The results, listed in Table~\ref{tab:cross-model}, show that the policy trained with ABUS excels only in ABUS evaluations, while the GenTUS-trained policy outperforms others in GenTUS and TUS evaluations, but performs slightly worse than ABUS-trained policies in ABUS evaluations. This result highlights the importance of cross-US training and evaluation to show the generalizability of the dialogue policy. Conducting such experiments is made straightforward in ConvLab-3, as the user simulator model can be easily changed within the configuration file.

\section{Conclusion}

In this paper, we present the dialogue system toolkit ConvLab-3, which puts a large number of datasets under one umbrella through our proposed unified data format. The usage of the unified format facilitates comparability and significantly reduces the implementation cost required for conducting experiments on multiple datasets. In addition, we provide recent powerful models for all components of a dialogue system and provide a convenient RL toolkit which enables researchers to easily build, train, analyze and evaluate dialogue systems.

We showcase the advantages of the unified format and RL toolkit in a large number of experiments, ranging from pre-training to RL training. The release of ConvLab-3 supports the community in developing the next generation of task-oriented dialogue systems.

\section{Limitations}

As ConvLab-3 is built for text-based TOD systems, we do not currently provide support for speech. One solution for this is the usage of speech recognition and text-to-speech interfaces such as Whisper \cite{radford23-whisper} and WaveNet \cite{wavenet16}. Secondly, while we provide several datasets in the unified format together with conversion scripts, the conversion of a new dataset still requires manual effort such as normalizing ontologies and transforming dialogue annotations. Lastly, ConvLab-3, currently, only supports the commonly used hierarchical dialogue state representation~\citep{budzianowski-etal-2018-multiwoz} but not yet state representations such as the graph-based state~\citep{andreas-etal-2020-task} and tree-structure state~\citep{cheng-etal-2020-conversational}. We consider these limitations as future work to further improve our toolkit.

\section*{Acknowledgements}
This work was supported by the National Key Research and Development Program of China (No. 2021ZD0113304), the National Science Foundation for Distinguished Young Scholars (with No. 62125604), the NSFC projects (Key project with No. 61936010 and regular project with No. 61876096), and sponsored by Tsinghua-Toyota Joint Research Fund.

This work was also supported by the DYMO project which has received funding from the European Research Council (ERC) provided under the Horizon 2020 research and innovation programme (Grant agreement No. STG2018 804636). In addition, it was supported by an Alexander von Humboldt Sofja Kovalevskaja Award endowed by the German Federal Ministry of Education and Research. Computational infrastructure and support were provided by the Centre for Information and Media Technology at Heinrich Heine University Düsseldorf and the Google Cloud Platform.

\bibliography{anthology,custom}
\bibliographystyle{acl_natbib}

\appendix

\section{Annotations of current unified datasets}
\label{sec:annotation} 

The statistics and annotations of already supported datasets are listed in Table~\ref{tab:datasets}.

\begin{table*}[t]
\small
\centering
\setlength{\tabcolsep}{0.9mm}{
\begin{tabular}{c|ccc|cccccc}
\toprule
\multirow{2}{*}{Dataset} & \multicolumn{3}{c|}{Statistics} & \multicolumn{6}{c}{Dataset Annotations} \\
& \#Dialogues & Avg. Turns & Domains & Goal & DA-U & DA-S & State & API Result & Database \\
\midrule
Camrest \cite{wen-etal-2016-conditional} & 676 & 10.8 & 1 & \checkmark & \checkmark & \checkmark & \checkmark & & \checkmark \\
WOZ 2.0 \cite{mrksic-etal-2017-neural} & 1200 & 7.4 & 1 & & \checkmark & & \checkmark & & \\
KVRET \cite{eric-etal-2017-key} & 3030 & 5.3 & 3 & & \checkmark & & \checkmark & \checkmark \\
DailyDialog \cite{li-etal-2017-dailydialog} & 13118 & 7.9 & 10 & & \checkmark & & & & \\
Taskmaster-1 \cite{byrne-etal-2019-taskmaster} & 13175 & 21.2 & 6 & & \checkmark & \checkmark & \checkmark & & \\
Taskmaster-2 \cite{byrne-etal-2019-taskmaster} & 17303 & 16.9 & 7 & & \checkmark & \checkmark & \checkmark & & \\
MultiWOZ 2.1 \cite{eric-etal-2020-multiwoz} & 10438 & 13.7 & 8 & \checkmark & \checkmark & \checkmark & \checkmark & & \checkmark \\
Schema-Guided \cite{rastogi2020sgd} & 22825 & 20.3 & 45 & & \checkmark & \checkmark & \checkmark & \checkmark & \\
MetaLWOZ \cite{li2020metalwoz} & 40203 & 10.4 & 51 &  \checkmark & & & & & \\
CrossWOZ \cite{zhu-etal-2020-crosswoz} & 6012 & 16.9 & 6 & \checkmark & \checkmark & \checkmark & \checkmark & \checkmark & \checkmark \\
Taskmaster-3 \cite{byrne-etal-2021-tickettalk} & 23757 & 20.1 & 1 & & \checkmark & \checkmark & \checkmark & \checkmark & \\
EmoWOZ \cite{feng-etal-2022-emowoz} & 11434 & 14.6 & 8 & \checkmark & \checkmark & \checkmark & \checkmark & & \checkmark \\
\bottomrule
\end{tabular}
}
\caption{Statistics and annotations of current unified datasets. DA-U/DA-S is dialogue acts annotation of user/system.}
\label{tab:datasets}
\end{table*}

\section{Tasks and models supported by the unified data format}
\label{sec:tasks}
Tasks and models already supported by the unified data format are shown in Table~\ref{tab:tasks}.
\begin{table*}[h]
\small
\centering
\setlength{\tabcolsep}{5mm}{
\begin{tabular}{c|c|c|c}
\toprule
Task & Input & Output & Models \\
\midrule
RG              & Context               & Response          & T5RG, LLMs             \\
Goal2Dial       & Goal                  & Dialogue          & T5Goal2Dialogue             \\
NLU             & Context               & DA-U              & T5NLU, BERTNLU, MILU, LLMs     \\
DST             & Context               & State             & T5DST, (Set)SUMBT, TripPy, LLMs   \\
Policy          & State, DA-U, Database & DA-S              & DDPT, PPO, PG         \\
Word-Policy          & Context, State, Database & Response              & LAVA       \\
NLG             & DA-S                  & Response          & T5NLG, SC-GPT, LLMs         \\
End2End         & Context, Database     & State, Response   & SOLOIST               \\
User Simulator  & Goal, DA-S            & DA-U, (Response)     & TUS, GenTUS, EmoUS, LLMs \\ 
\bottomrule
\end{tabular}
}

\caption{Tasks and models supported by the unified data format. RG is response generation without database support. Goal2Dial is generating a dialogue from a user goal. NLU is natural language understanding. BERTNLU, MILU, PPO, PG are from ConvLab-2 \cite{zhu-etal-2020-convlab}. Currently supported LLMs include LLaMA-2 \cite{llama2}, ChatGLM2 \cite{zeng2022glm}, and OpenAI models such as ChatGPT (through API).}
\label{tab:tasks}
\end{table*}

\section{Example Serialized Dialogue Acts and State} \label{appendix:section:serialization}
The example of serialized dialogue acts and states is shown in Table~\ref{tab:serialization}.
\begin{table}[h]
\centering
\small
\begin{tabular}{@{}p{7.7cm}@{}}
\toprule
  \textbf{User}: I am looking for a \textbf{\textit{cheap}} restaurant. \\
  \textbf{System}: Is there a particular area of town you prefer? \\
  \textbf{User}: In the \textbf{\textit{centre}} of town. \\ \midrule
  \textbf{DA-U}:[inform][restaurant]([area][centre]) \\
  \textbf{State}: [restaurant]([area][centre],[price range][cheap]) \\
  \textbf{DA-S}: [recommend][restaurant]([name][Zizzi Cambridge]) \\ \midrule
  \textbf{System}: I would recommend \textbf{\textit{Zizzi Cambridge}}. \\
\bottomrule
\end{tabular}
\caption{Example serialized dialogue acts and state. 
Dialogue acts are in the form of ``[intent] [domain] ([slot] [value],...);...''. State is in the form of ``[domain] ([slot] [value],...);...''. Multiple items are separated by a semicolon.}
\label{tab:serialization}
\end{table}

\section{Joint Training}
\label{sec:app_joint_training}

In this experiment, we investigate the effect of training a model on multiple datasets jointly instead of separately.
For joint training, we merge MultiWOZ, SGD, and Taskmaster datasets into one and train a single model, which requires the model to handle datasets with different ontologies.
Intuitively, the advantage of joint training is that knowledge transfer is bi-directional and persists for the whole training period, while the disadvantage is that there may be inconsistent labels for similar inputs on different datasets, potentially confusing the models.

To avoid confusion, for T5-NLU, T5-DST, and T5-NLG, we prepend the dataset name to the original input to distinguish data from different datasets.
For SetSUMBT, we only predict the state of the target dataset.
Since SGD may have several services for one domain, we normalize the service name to the domain name (e.g., \texttt{Restaurant\_1} to \texttt{Restaurant}) when evaluating NLU and DST.
However, similar slots of different services (e.g., \texttt{city} and \texttt{location}) will still confuse the model.
While further normalization may help, we are aiming to compare independent training and joint training instead of achieving SOTA performance.
For Taskmaster-1/2/3, we evaluate each sample with the corresponding ontology and then calculate the metrics on all test samples of three datasets.
In addition, on SGD and Taskmaster, we build pseudo user goals for TUS and GenTUS by accumulating constraints and requirements in user dialogue acts during conversations.

We compare independent training and joint training in Table \ref{tab:multitask}.
MultiWOZ, SGD, and Taskmaster have 8K, 16K, and 43K dialogues for training respectively.
Joint training on these datasets does not lead to substantial performance drops in most cases, indicating that models have sufficient capacity to encode knowledge of different datasets simultaneously.
However, joint training does not always improve performance either.
It consistently improves the End2End model SOLOIST but makes no difference to T5-NLU.
For other models, the gains vary with the dataset.
Associating with the previous pre-training-then-fine-tuning experiment, we think the difference may be attributed to the varying task complexity on different datasets.
When the original data of a certain dataset are sufficient for a model to solve the task, including other datasets via joint training may not bring further benefit.

\begin{table}[h]
\small
\centering
\setlength{\tabcolsep}{0.5mm}{
\begin{tabular}{@{}lcc|cc|cc@{}}
\toprule
                        & \multicolumn{2}{c}{MultiWOZ 2.1} & \multicolumn{2}{c}{SGD} & \multicolumn{2}{c}{Taskmaster}\\ \cmidrule{2-7} 
\textbf{NLU}            & ACC       & F1        & ACC       & F1        & ACC       & F1                \\ \midrule
\multirow{2}{*}{T5-NLU} & 77.8      & 86.5      & 45.0      & 58.6      & 81.8      & 73.0              \\
                        & 77.5      & 86.4      & 45.2      & 58.6      & 81.8      & 73.0              \\ \midrule \midrule
\textbf{DST}            & JGA       & Slot F1   & JGA       & Slot F1   & JGA       & Slot F1           \\ \midrule
\multirow{2}{*}{T5-DST} & 52.6      & 91.9      & 20.1      & 58.5      & 48.5      & 81.1              \\
                        & 53.1      & 91.9      & 20.6      & 60.0      & 48.6      & 81.0              \\ \midrule
\multirow{2}{*}{SetSUMBT}& 50.3         & 90.8          & 20.0          & 58.8          & 24.9          & 65.5                  \\
                        & 50.8          & 91.0          & 21.1	 & 59.2          & 25.3          & 67.0                  \\ \midrule \midrule
\textbf{NLG}            & SER $\downarrow$      & BLEU      & SER $\downarrow$      & BLEU      & SER $\downarrow$      & BLEU              \\ \midrule
\multirow{2}{*}{T5-NLG} & 3.7       & 35.8      & 11.9      & 29.6      & 2.1       & 51.5              \\
                        & 3.2       & 35.6      & 8.3       & 29.9      & 2.0       & 51.3              \\ \midrule
\multirow{2}{*}{SC-GPT} &      4.8     &  33.6         &       9.6    &   28.2        &      2.2     &   47.9                \\
                        &      3.3     &       33.5    &    6.8       &           29.8 &       1.6    &    47.3               \\ \midrule \midrule
\textbf{End2End}        & Comb.     & BLEU      & Slot F1     & BLEU      & Slot F1     & BLEU              \\ \midrule
\multirow{2}{*}{SOLOIST}& 67.0      & 16.8      &    56.9       &    11.2       &    8.5       &  28.0                 \\
                        & 71.4      & 17.1      &     69.7      &  23.1         &      9.2     &   29.2                \\ \midrule \midrule
\textbf{US-DA}          & ACC       & F1        & ACC       & F1        & ACC       & F1                \\ \midrule
\multirow{2}{*}{TUS}    & 15.0      & 53.2      & 10.2      & 11.6      & 23.0      & 23.0              \\
                        & 32.0      & 62.3      & 13.8      & 15.6      & 22.5      & 22.0              \\ \midrule
\textbf{US-NL}          & SER $\downarrow$      & F1        & SER $\downarrow$      & F1        & SER $\downarrow$      & F1                \\ \midrule
\multirow{2}{*}{GenTUS} & 4.2       & 62.5      & 8.4       & 48.8      & 4.2       & 43.8              \\
                        & 3.3       & 49.5      & 8.0       & 48.7      & 3.5       & 43.9              \\ \bottomrule
\end{tabular}
}
\caption{Comparison of independent training and joint training (1st row vs. 2nd row of each model) on 3 datasets. We normalize the service name to the domain name when evaluating NLU and DST on SGD.}
\vspace{-1em}
\label{tab:multitask}
\end{table}

\section{Retrieval Augmentation}
\label{app:retrieval-augmentation}

We further explore transferring knowledge from other datasets through retrieval-based data augmentation.
Here we only consider the single-turn NLU task where the input is an utterance since utterance-level similarity is easier to model than dialogue-level similarity.
For each utterance in the target dataset, we retrieve the top-k (k $\in \{1,3\}$) most similar utterances from other datasets measured by the MiniLMv2 model \cite{wang-etal-2021-minilmv2} using Sentence Transformers \cite{reimers-gurevych-2019-sentence}.
We then use retrieved samples in two ways:
\begin{enumerate}
    \item Augment training data. Models are trained on both original training data and retrieved data. 
    \item Additionally input the retrieved samples as in-context examples, including retrieved utterances and their dialogue acts, as shown in Table \ref{tab:in-context}. Different from \citet{liu-etal-2022-makes}, the retrieved samples are from other datasets instead of the target dataset and we will train models on augmented samples.
\end{enumerate}

Since different datasets have different ontologies (i.e., definitions of intent, domain, slot), we prepend the corresponding dataset name to an input utterance as in the joint training experiment.
We use the T5-NLU model and try two model sizes \texttt{T5-Small} and \texttt{T5-Large}.
We fine-tune the models on MultiWOZ using the same settings as in the pre-training-then-fine-tuning experiment.

\begin{table}[h]
\centering
\small
\begin{tabular}{@{}p{7.7cm}@{}}
\toprule
  \textbf{Original input}: user: Yes please, for 8 people at 18:30 on thursday. \\ \midrule
  \textbf{Augmented input}: \colorbox{yellow}{\textbf{tm3}} user: Yes, please, four for 8:10pm. \\=> [inform] [movie]([num.tickets][four],[time.showing][8:10 pm]) \colorbox{brown}{\textbf{tm1}} user: Yes, 8PM, please. => [inform][restaurant \_reservation]([time.reservation][8PM]) \colorbox{cyan}{\textbf{sgd}} user: Yes please, for 3 people on March 8th at 12:30 pm. => [affirm\_intent] [Restaurants\_1]([][]);[inform][Restaurants\_1] ([party\_size][3]\\
  {[date][March 8th],[time][12:30 pm])} \colorbox{green}{\textbf{multiwoz21}} user: Yes please, for 8 people at 18:30 on thursday. \\ \midrule
  \textbf{Output}: [inform][restaurant]([book day][thursday],[book time][18:30],[book people][8]) \\
\bottomrule
\end{tabular}
\caption{An example of input augmented by retrieved top-3 samples from other TOD datasets for in-context learning. Dataset names are highlighted.}

\label{tab:in-context}
\end{table}

\begin{table}[h]
\small
\centering
\begin{tabular}{@{}lcc|cc|cc@{}}
\toprule
\multirow{2}{*}{\textbf{T5-NLU}}                    & \multicolumn{6}{c}{\qquad \ MultiWOZ 2.1}                                         \\
                    & \multicolumn{2}{c}{1\%} & \multicolumn{2}{c}{10\%} & \multicolumn{2}{c}{100\%}  \\ \cmidrule{2-7} 
\textbf{T5-small}   & ACC         & F1          & ACC         & F1          & ACC       & F1  \\ \midrule
Baseline       & 48.1  & 64.6  & 68.8  & 80.6  & 77.8  & 86.5\\
Pre-trained         & 55.5  & 70.1  & 69.8  & 81.0  & 77.9    & 86.5\\ \midrule
\textit{Data Aug.}  &             &             &             &             &           &       \\
\quad - top1        & 51.4  & 66.8  & 69.0  & 80.6  & 77.5    & 86.5\\
\quad - top3        & 51.3  & 66.4  & 68.8  & 80.6  & 77.0    & 86.2\\ \midrule
\textit{In-context} &             &             &             &             &           &       \\
\quad - top1      & 44.5  & 61.3  & 68.5  & 80.1  & 77.7    & 86.4\\
\quad - top3    & 43.6  & 60.8  & 68.1  & 79.8  & 77.5    & 86.4\\ \midrule \midrule
\textbf{T5-large}   & ACC         & F1          & ACC         & F1          & ACC       & F1  \\ \midrule
Baseline       & 51.2  & 67.9  & 67.8  & 80.0  & 76.8  & 86.4\\
Pre-trained         & 56.7  & 71.7  & 69.5  & 81.0  & 76.8  & 86.1\\ \midrule
\textit{Data Aug.}  &             &             &             &             &           &       \\
\quad - top1        & 49.7  & 66.8  & 68.7  & 80.6  & 76.9  & 86.1\\
\quad - top3        & 48.5  & 66.2  & 68.5  & 80.5  & 76.3  & 85.8\\ \midrule
\textit{In-context} &             &             &             &             &           &       \\
\quad - top1      & 43.4  & 61.3  & 69.1  & 81.0  & 76.5  & 85.9\\
\quad - top3    & 43.9  & 63.0  & 68.7  & 80.8  & 76.9  & 86.2\\ \bottomrule
\end{tabular}
\caption{Comparison of different ways to use other TOD datasets: (1) pre-training, (2) retrieving similar samples for data augmentation or (3) in-context learning.}
\label{tab:retrieval}
\end{table}

\section{RL Experiments}
\label{app:rl-experiments-chris}

\subsection{Transfer Learning} 

\begin{figure*}[t!]
  \begin{subfigure}[t]{0.32\textwidth}
    \includegraphics[scale=0.35]{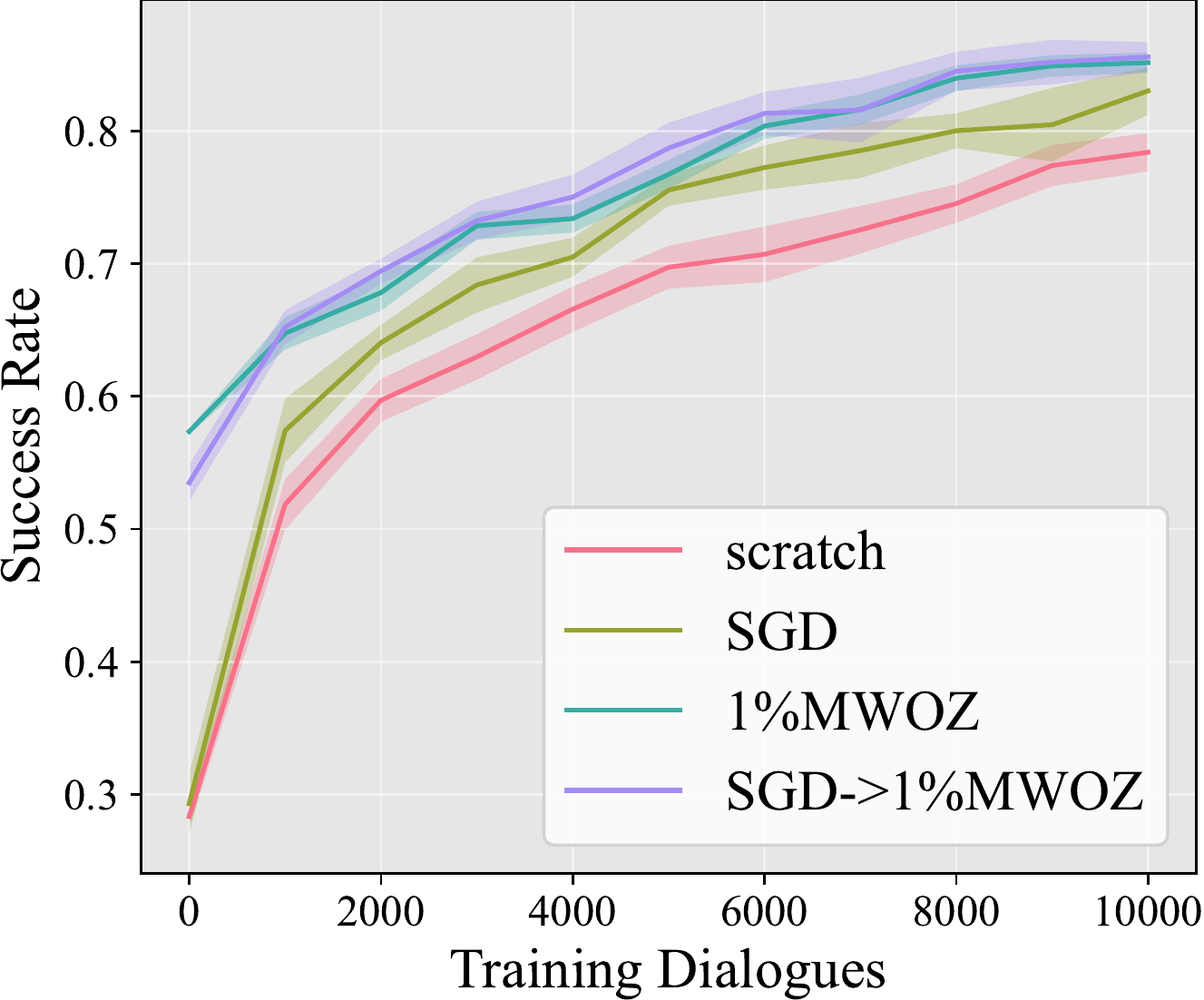}
    \caption{Success rate}
  \end{subfigure}
  \hspace{0.01\textwidth}%
  \begin{subfigure}[t]{0.32\textwidth}
    \includegraphics[scale=0.35]{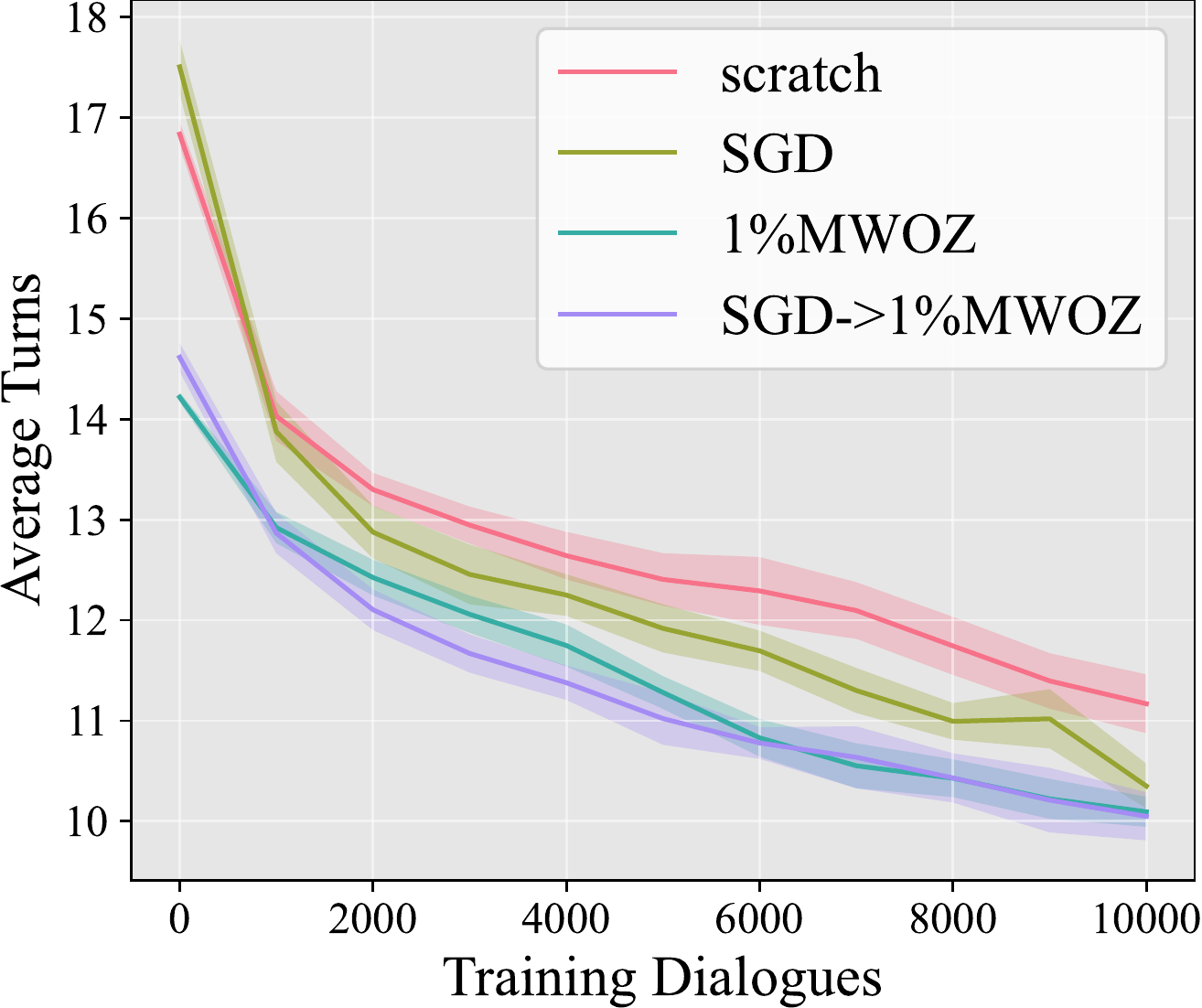}
    \caption{Average number of turns}
  \end{subfigure}
  \hspace{0.01\textwidth}%
  \begin{subfigure}[t]{0.32\textwidth}
    \includegraphics[scale=0.35]{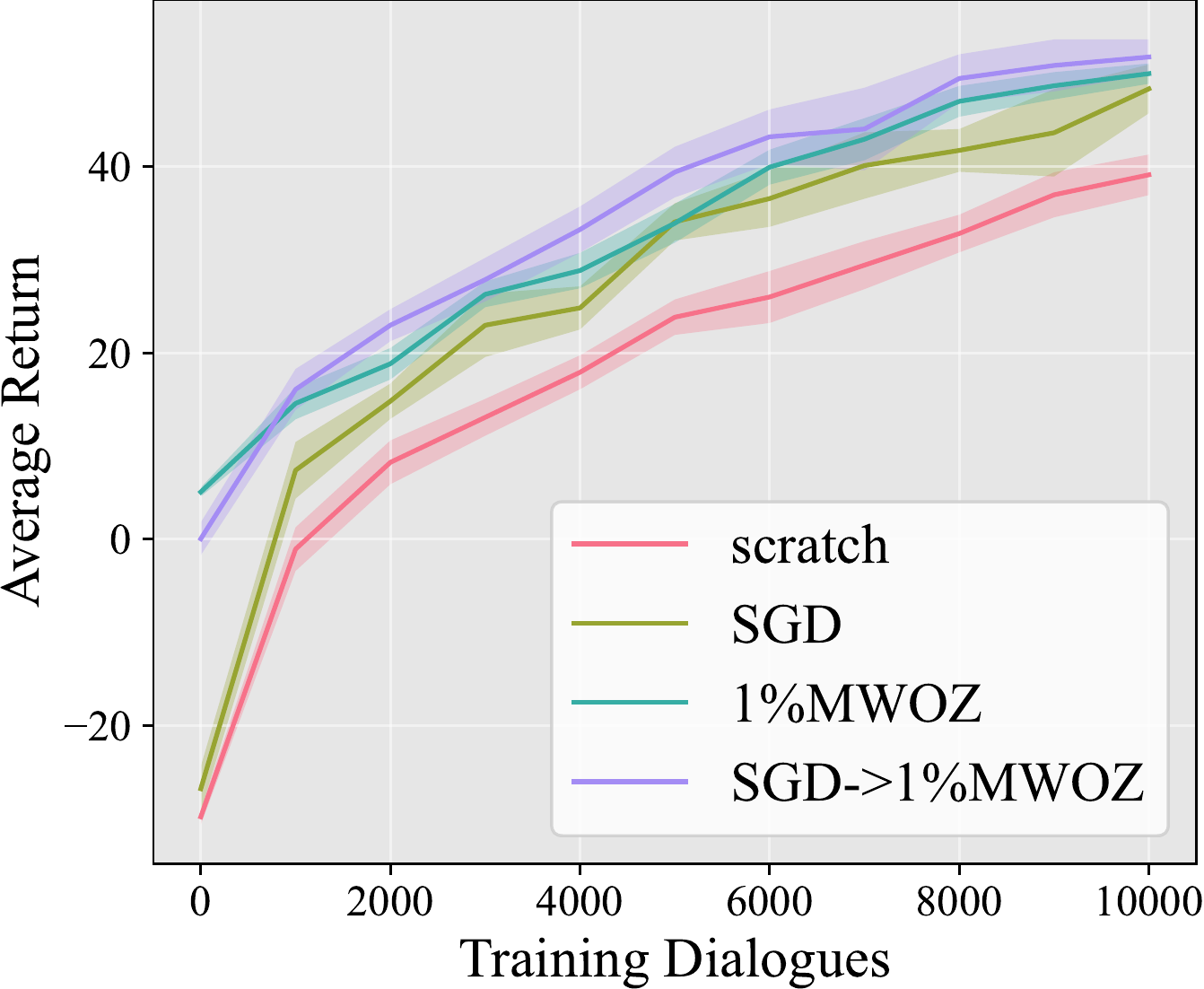}
    \caption{Average return per episode}
  \end{subfigure}

    \begin{subfigure}[t]{0.32\textwidth}
    \includegraphics[scale=0.35]{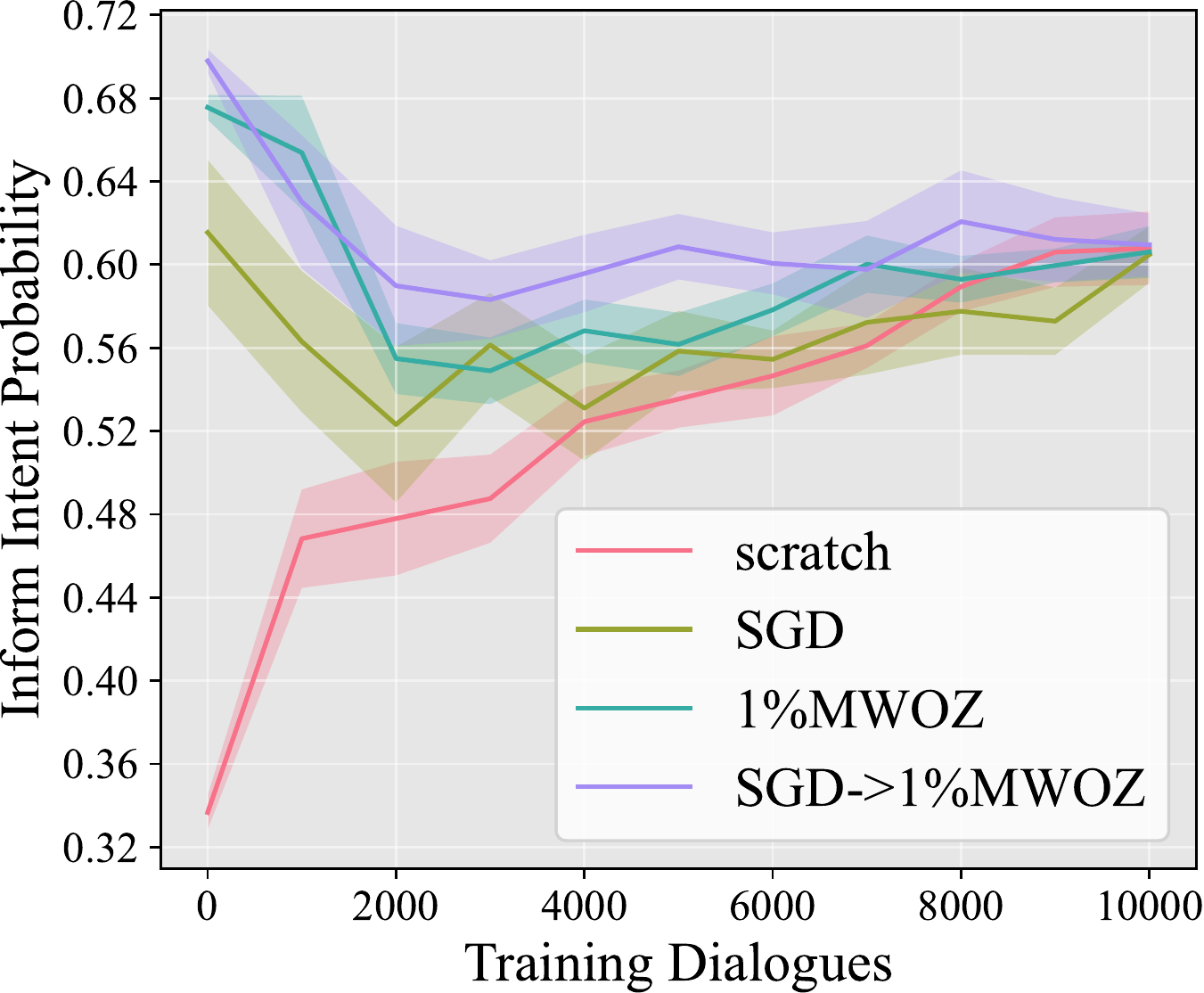}
    \caption{Probability of the inform intent}
  \end{subfigure}
  \hspace{0.01\textwidth}%
  \begin{subfigure}[t]{0.32\textwidth}
    \includegraphics[scale=0.35]{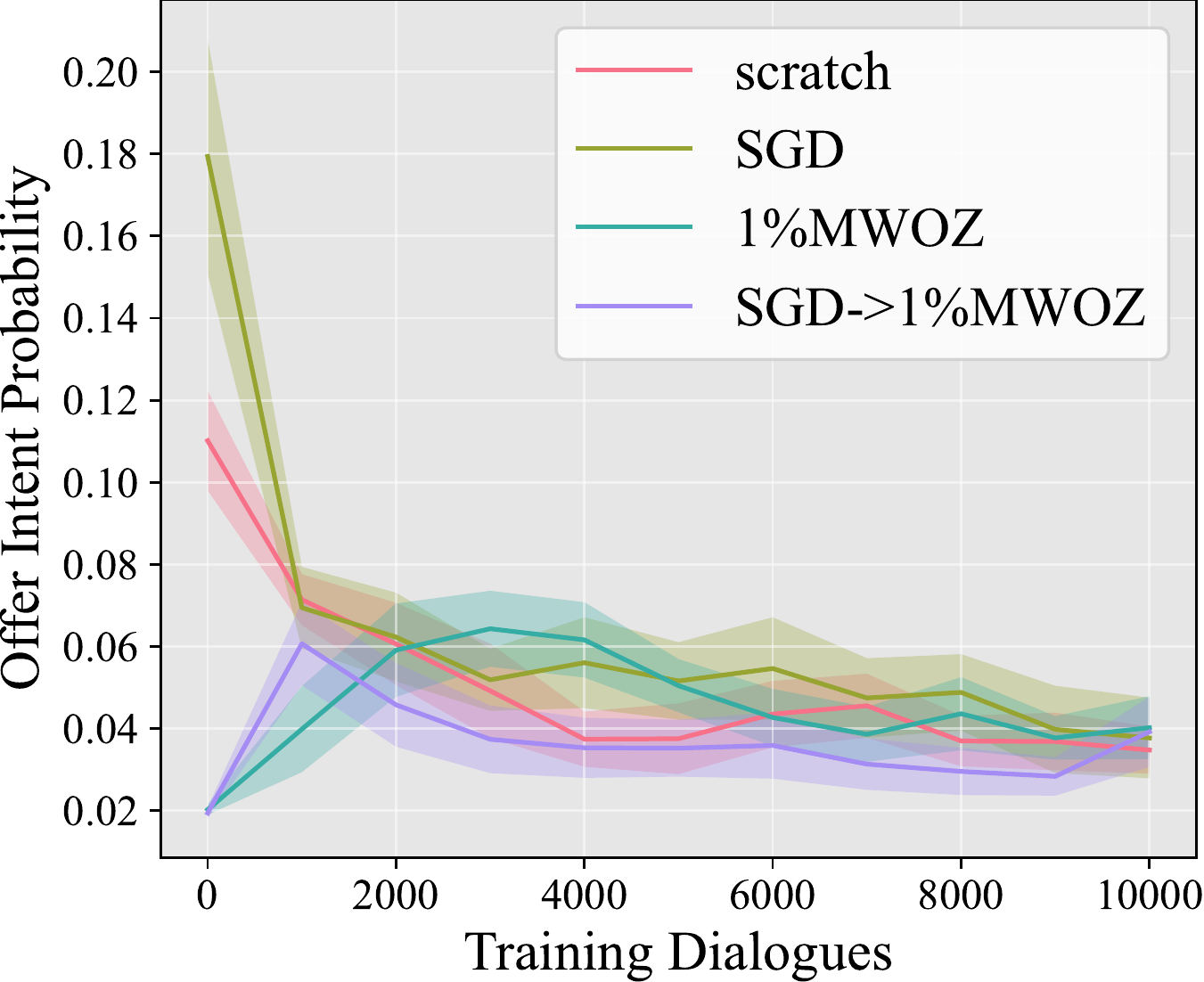}
    \caption{Probability of the offer intent}
  \end{subfigure}
  \hspace{0.01\textwidth}%
  \begin{subfigure}[t]{0.32\textwidth}
    \includegraphics[scale=0.35]{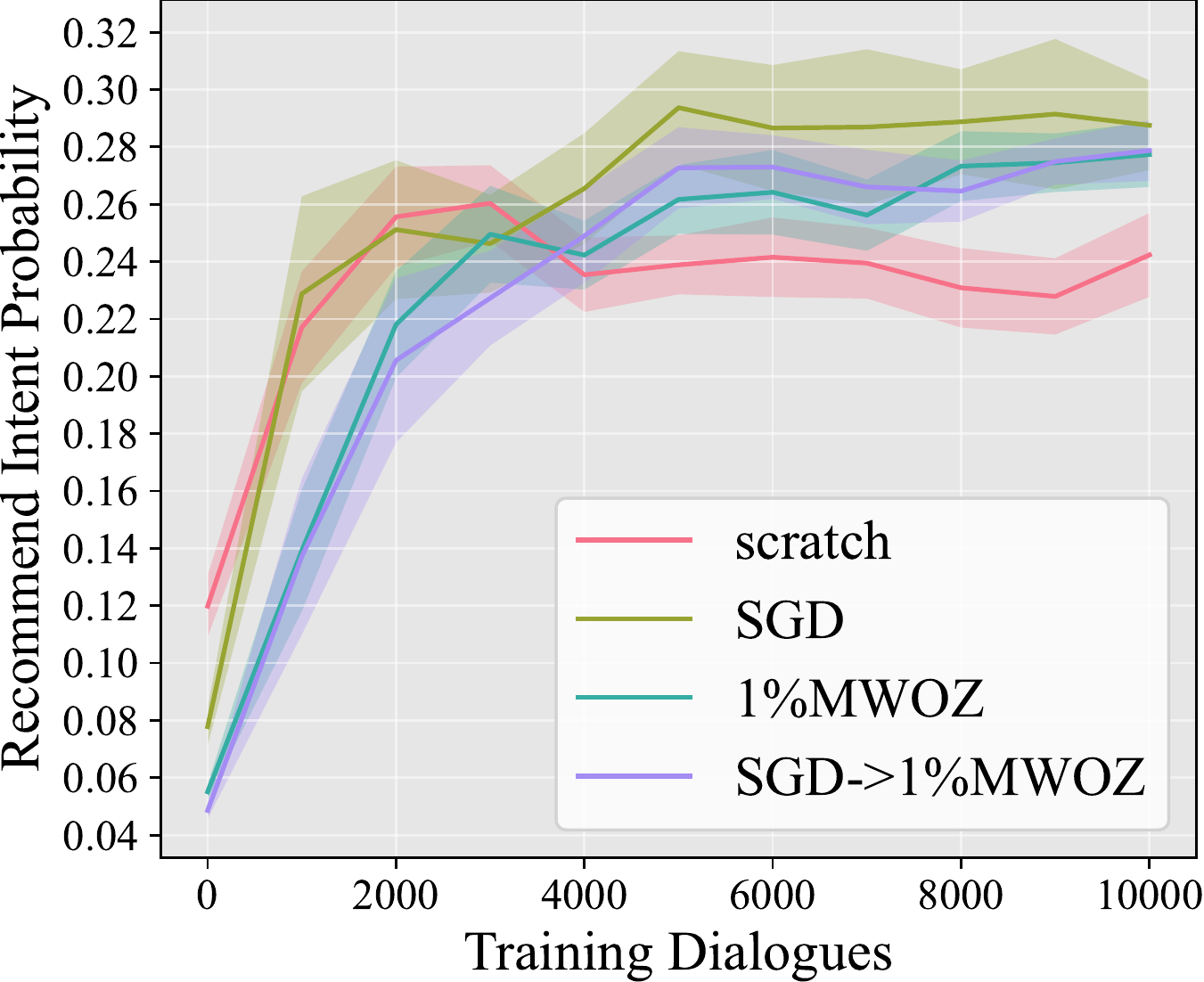}
    \caption{Probability of the recommend intent}
  \end{subfigure}

\caption{Pre-training then RL training experiments with the DDPT model in interaction with the rule-based simulator. Shaded regions show standard error. Each model is evaluated on 9 different seeds.}
\label{fig:app-pretrain-rl}
\end{figure*}

In this section, we provide additional plots for the experiments conducted in Section \ref{sec:rl-experiments}. Figure \ref{fig:app-pretrain-rl} depicts success rate (which is less strict success rate), average number of turns as well as the average return. Moreover, we show additional intent probabilities. We can observe in Figure \ref{fig:app-pretrain-rl}(e) that the policy pre-trained on only SGD data uses more offer intents intially, which reflects the behaviour in the data set. The probability of the offer intent then decreases whereas the probability for the recommend intent increases during learning.

\begin{figure*}[t!]
  \begin{subfigure}[t]{0.32\textwidth}
  \centering
    \includegraphics[scale=0.35]{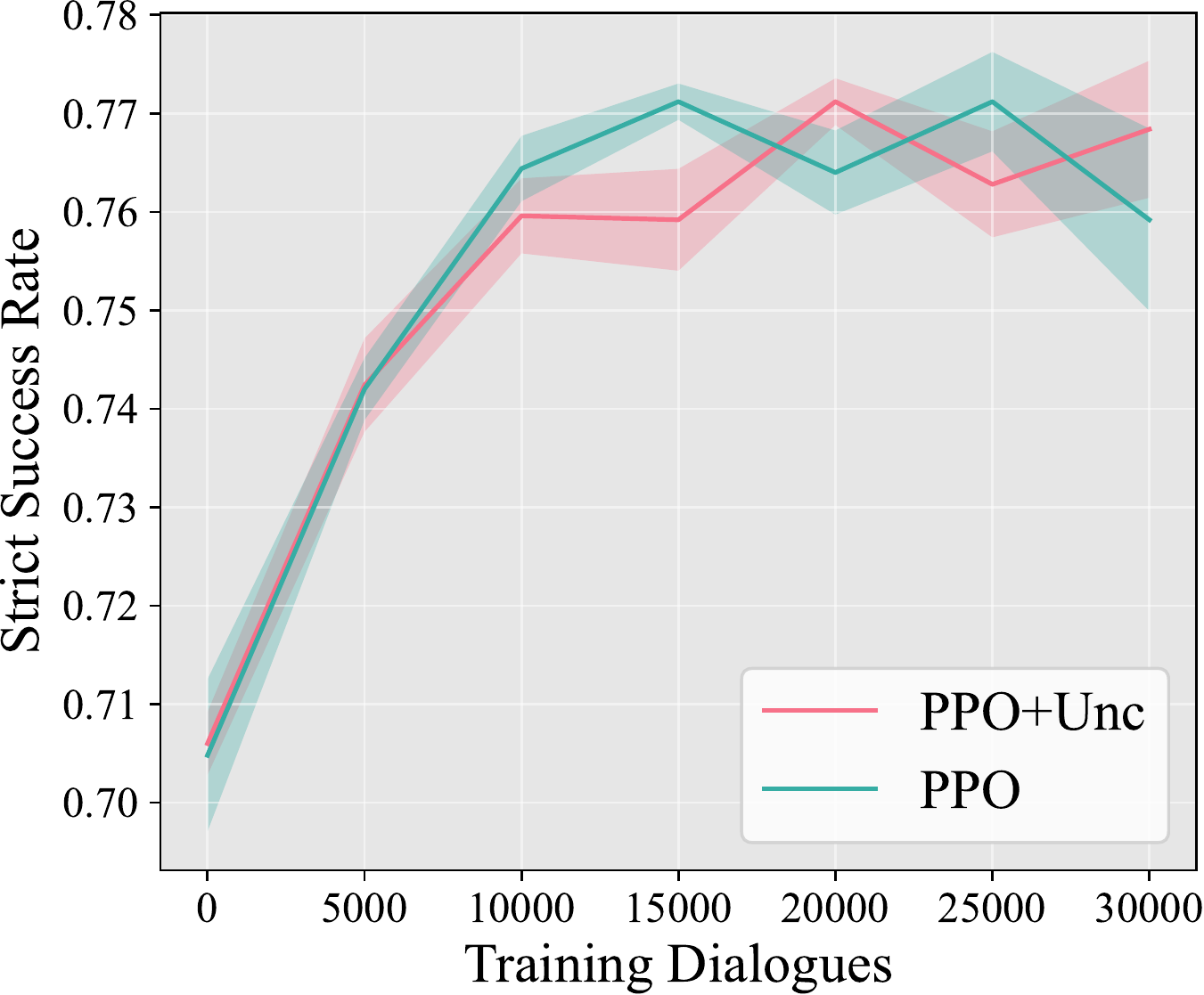}
    \caption{Strict success rate}
  \end{subfigure}
  \hspace{0.01\textwidth}%
  \begin{subfigure}[t]{0.32\textwidth}
  \centering
    \includegraphics[scale=0.35]{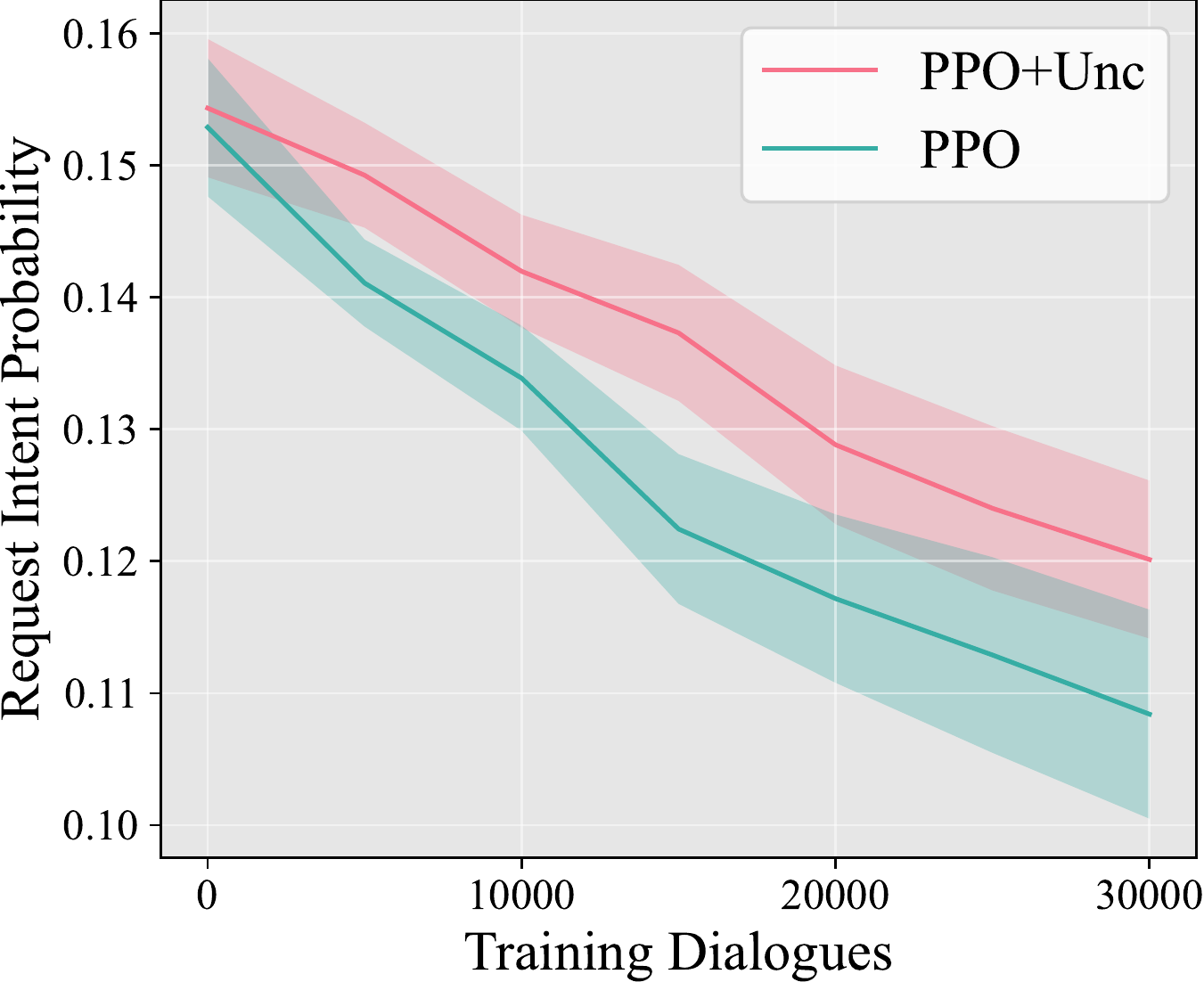}
    \caption{Probability of utilising the \textit{Request} intent in a turn.}
  \end{subfigure}
  \hspace{0.01\textwidth}%
  \begin{subfigure}[t]{0.32\textwidth}
  \centering
    \includegraphics[scale=0.35]{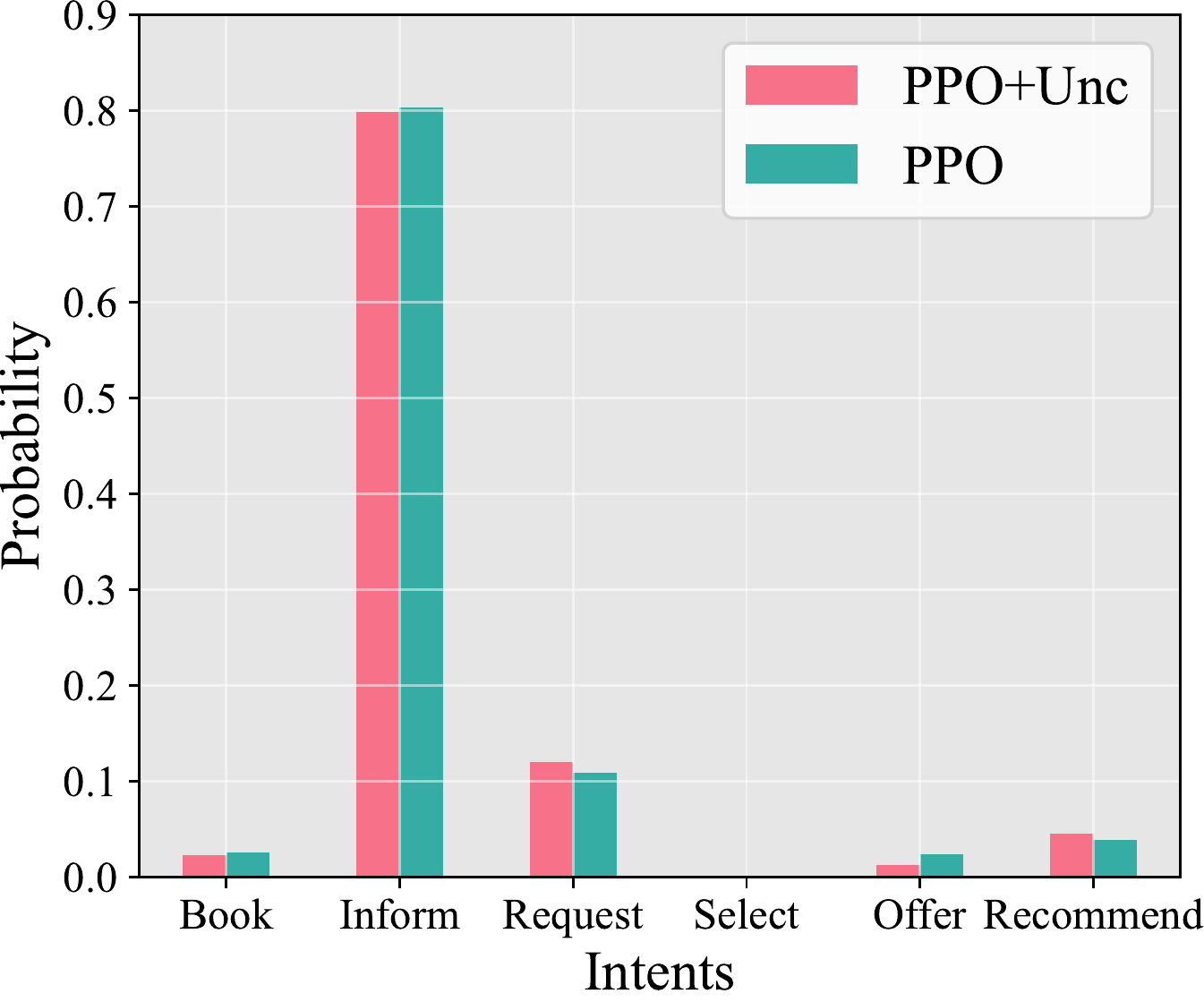}
    \caption{Distribution of intents for the final policies.}
  \end{subfigure}

\caption{Evaluation of the PPO policy trained combined with a SetSUMBT DST model with and without uncertainty features respectively. The policy is trained in an environment that contains $5\%$ user NLG noise to illustrate the impact of uncertainty.}
\label{rl-uncertainty}
\end{figure*}

\subsection{Training with Uncertainty Features} \label{app:uncertainty-features}

\subsubsection{Experimental Setup}
The simplified vectorizer module in ConvLab-3 makes it possible to easily include dialogue related features that might benefit dialogue policy learning. One such example is given due to the problem of resolving ambiguities in conversations. Humans naturally identify these ambiguities and resolve the uncertainty resulting from them. For a dialogue system to be robust to ambiguity, it is crucial to identify and resolve these uncertainties \cite{van-niekerk-etal-2021-uncertainty}.

ConvLab-3 provides the SetSUMBT dialogue belief tracker which achieves SOTA performance in terms of the accuracy of its uncertainty estimates. Using the vectorizer class to incorporate these features we can train a policy using the uncertainty features obtained from SetSUMBT. To illustrate the effectiveness of uncertainty features during RL, we train a PPO policy using these features. The template-based NLG in ConvLab-3 also allows for the inclusion of noise~\cite{van-niekerk-etal-2021-uncertainty} in generated responses which allows for uncertainties to arise during the conversation, simulating a more realistic conversation.

\subsubsection{Result Analysis}

Figure \ref{rl-uncertainty} (a) reveals that the policy trained using uncertainty features performs at least as well as the policy trained without these features. \citet{van-niekerk-etal-2021-uncertainty} further showed that the policy trained with uncertainty features performs significantly better in conversation with humans than the policy trained without. This is an indication that the policy using uncertainty features can handle ambiguities in conversation better than the policy without uncertainty modeling. To investigate how this policy resolves uncertainty we analyze the action distributions of the policy using the new RL toolkit evaluation tools, which provide new insights into the behaviour of dialogue policy modules. Figure \ref{rl-uncertainty} (b) and (c) show that the policy trained using uncertainty features utilizes significantly more request actions than the policy without these features. This indicates that the policy aims to resolve uncertainty by requesting information from the user. For instance, if the policy recognizes uncertainty regarding the price range a user has requested, it can resolve this through the use of a request. See \citet{van-niekerk-etal-2021-uncertainty} for example dialogues with humans where this can be observed.

\section{Configuring the Dialogue Pipeline and User Environment} \label{appendix:section:config-file}

The modules in the dialogue system pipeline and the user environment for interaction are specified using a configuration file as can be seen in Listing \ref{listing:configuration-file}.

\section{Ontology, Data and Database in the Unified Format}
\label{appendix:section:multiwoz-example}

As explained in Section \ref{unified-data-format-section}, a dataset in the unified format consists of an ontology, dialogues and a database. We depict an example for ontology, dialogues and database results for the MultiWOZ 2.1 dataset in the unified format in Listing \ref{listing:ontology-example}, \ref{listing:data-example} and \ref{listing:database-result}, respectively.

\begin{figure*}[t!]
\begin{lstlisting}[language=Python, numbers=none, caption={Example configuration file for a PPO dialogue policy, binary vectoriser, BertNLU and rule-based DST. The user simulator is GenTUS. We can straightforwardly build different configurations by substituting different modules.}, label={listing:configuration-file}] 
{
    "model": {
        "load_path": "from_pretrained",
        "pretrained_load_path": "",
        "use_pretrained_initialisation": false,
        "batchsz": 200,
        "seed": 0,
        "epoch": 100,
        "eval_frequency": 5,
        "process_num": 1,
        "num_eval_dialogues": 20,
        "sys_semantic_to_usr": false
    },
    "vectorizer_sys": {
        "uncertainty_vector_mul": {
            "class_path": "convlab.policy.vector.vector_binary.VectorBinary",
            "ini_params": {
                "use_masking": true,
                "manually_add_entity_names": true,
                "seed": 0
            }
        }
    },
    "nlu_sys": {
        "BertNLU": {
            "class_path": "convlab.nlu.jointBERT.unified_datasets.BERTNLU",
            "ini_params": {
                "mode": "all",
                "config_file": "multiwoz21_all.json",
                "model_file": "https://huggingface.co/ConvLab/bert-base-nlu/resolve/main/bertnlu_unified_multiwoz21_all_context0.zip"
            }
        }
    },
    "dst_sys": {
        "RuleDST": {
            "class_path": "convlab.dst.rule.multiwoz.dst.RuleDST",
            "ini_params": {}
        }
    },
    "sys_nlg": {},
    "nlu_usr": {},
    "dst_usr": {},
    "policy_usr": {
        "GenTUS": {
            "class_path": "convlab.policy.genTUS.stepGenTUS.UserPolicy",
            "ini_params": {
                "model_checkpoint": "convlab/policy/genTUS/unify/experiments/multiwoz21-exp",
                "mode": "language",
                "only_action": false
            }
        }
    },
    "usr_nlg": {}
}

\end{lstlisting}
\end{figure*}

\begin{figure*}[t!]
\begin{lstlisting}[language=Python, numbers=none, caption={Example of the unified format ontology for the MultiWOZ $2.1$ dataset.}, label={listing:ontology-example}]
{
    "domains": {
        "attraction": {
            "description": "find an attraction",
            "slots": {
                "area": {
                    "description": "area to search for attractions",
                    "is_categorical": true,
                    "possible_values": ["centre", "east", "north", "south", "west"]
                    },
                "name": {
                    "description": "name of the attraction",
                    "is_categorical": false,
                    "possible_values": []
                },
                ...
            }
        },
        ...
    },
    "intents": {
        "inform": {"description": "inform the value of a slot"},
        "request": {"description": "ask for the value of a slot"},
        ...
    },
    "state": {
        "attraction": {
            "type": "",
            "name": "",
            "area": ""
        },
        "hotel": {
            "name": "",
            "area": "",
            ...
        },
        ...
    },
    "dialogue_acts": {
        "categorical": [
            "{'user': False, 'system': True, 'intent': 'nobook', 'domain': 'hotel', 'slot': 'book day'}",
            "{'user': False, 'system': True, 'intent': 'nobook', 'domain': 'restaurant', 'slot': 'book day'}",
            ...
        ],
        "non-categorical": [
            "{'user': False, 'system': True, 'intent': 'inform', 'domain': 'attraction', 'slot': 'address'}",
            "{'user': False, 'system': True, 'intent': 'inform', 'domain': 'attraction', 'slot': 'choice'}",
            ...
        ],
        "binary": [
            "{'user': False, 'system': True, 'intent': 'book', 'domain': 'attraction', 'slot': ''}",
            "{'user': False, 'system': True, 'intent': 'book', 'domain': 'hospital', 'slot': ''}",
            ...
        ]
    }
}
\end{lstlisting}
\end{figure*}

\begin{figure*}[t!]
\begin{lstlisting}[language=Python, numbers=none, caption={Example of the unified format data within the MultiWOZ $2.1$ dataset.}, label={listing:data-example}]
[
    {
        "dataset": "multiwoz21",
        "data_split": "train",
        "dialogue_id": "multiwoz21-train-0",
        "original_id": "SNG01856.json",
        "domains": ["hotel", "general"],
        "goal": {
            "description": "You are looking for a place to stay. The hotel should be in the cheap price range and should be in the type of hotel...",
            "inform": {
                "hotel": {
                    "type": "hotel",
                    "parking": "yes",
                    "price range": "cheap",
                    "internet": "yes",
                    "book stay": "3|2",
                    "book day": "tuesday",
                    "book people": "6"
                }
            },
            "request": {
                "hotel": {}
            }
        },
        "turns": [
            {
                "speaker": "user",
                "utterance": "am looking for a place to to stay that has cheap price range it should be in a type of hotel",
                "utt_idx": 0,
                "dialogue_acts": {
                    "categorical": [
                        {
                            "intent": "inform",
                            "domain": "hotel",
                            "slot": "price range",
                            "value": "cheap"
                        }
                    ],
                    "non-categorical": [
                        {
                            "intent": "inform",
                            "domain": "hotel",
                            "slot": "type",
                            "value": "hotel",
                            "start": 87,
                            "end": 92
                        }
                    ],
                    "binary": []
                },
                "state": {
                    "attraction": {
                        "type": "",
                        "name": "",
                        "area": ""
                    },
                    ...
                }
            },
            {
                "speaker": "system",
                "utterance": "Okay, do you have a specific area you want to stay in?",
                "utt_idx": 1,
                "dialogue_acts": {
                    "categorical": [],
                    "non-categorical": [],
                    "binary": [
                        {
                            "intent": "request",
                            "domain": "hotel",
                            "slot": "area"
                        }
                    ]
                },
                "booked": {
                    "taxi": [],
                    "restaurant": [],
                    ...
                }
            },
            ...
        ]
    },
    ...
]
\end{lstlisting}
\end{figure*}

\begin{figure*}[t!]
\begin{lstlisting}[language=Python, numbers=none, caption={Example of database query result when searching for a moderately priced hotel in the east from the MultiWOZ unified format database.}, label={listing:database-result}] 
[
    {
        "address": "124 tenison road",
        "area": "east",
        "internet": "yes",
        "parking": "no",
        "id": "0",
        "name": "a and b guest house",
        "phone": "01223315702",
        "postcode": "cb12dp",
        "price": {
        "double": "70",
        "family": "90",
        "single": "50"
        },
        "pricerange": "moderate",
        "stars": "4",
        "takesbookings": "yes",
        "type": "guesthouse",
        "Ref": "00000000"
    },
    ...
]
\end{lstlisting}
\end{figure*}

\end{document}